\documentclass{article}

\usepackage[preprint]{neurips_2019}

\usepackage[utf8]{inputenc} 
\usepackage[T1]{fontenc}    
\usepackage{hyperref}       
\usepackage{url}            
\usepackage{booktabs}       
\usepackage{amsfonts}       
\usepackage{nicefrac}       
\usepackage{microtype}      

\usepackage{graphicx}
\usepackage{amsmath,amssymb} 
\usepackage{xcolor}

\usepackage{times}
\usepackage{epsfig}
\usepackage{graphicx}
\usepackage{amsmath}
\usepackage{amssymb}
\usepackage{pifont} 
\usepackage{bbm}
\usepackage{empheq}
\usepackage{rotating}
\usepackage{amsthm}
\usepackage{gensymb} 

\usepackage{bm}
\usepackage{siunitx}
\usepackage{textcomp}
\usepackage{subcaption}

\usepackage{caption}
\usepackage{subcaption}
\usepackage{algorithm,multicol,lipsum}

\usepackage{wrapfig}

\usepackage[noend]{algpseudocode}
\usepackage{multirow}
\usepackage{makecell}
\usepackage{arydshln}
\usepackage{booktabs}
\usepackage{wrapfig}
\usepackage{listings}
\graphicspath{ {images/} }

\usepackage{xspace}
\makeatletter
\DeclareRobustCommand\onedot{\futurelet\@let@token\@onedot}
\def\@onedot{\ifx\@let@token.\else.\null\fi\xspace}

\def\ie{\emph{i.e}\onedot}

\def\wrt{w.r.t\onedot}

\newcommand{\mr}[1]{[\textcolor{red}{MR: {#1}}]}

\makeatother

\newcommand{\dataset}{\mathcal{D}}
\newcommand{\epsmem}{\mathcal{M}}
\newcommand{\param}{\theta}

\newcommand{\ip}{\mathbf{x}}
\newcommand{\ipspace}{\mathcal{X}}
\newcommand{\descspace}{\mathcal{T}}

\newcommand{\op}{\mathbf{y}}
\newcommand{\opspace}{\mathcal{Y}}

\newcommand{\SKIP}[1]{}

\newcommand{\mtask}{\textsc{multi-task}\xspace}

\newcommand{\van}{\textsc{finetune}\xspace}

\newcommand{\ewc}{\textsc{ewc}\xspace}

\newcommand{\ewcp}{\textsc{ewc++}\xspace}

\newcommand{\gem}{\textsc{gem}\xspace}
\newcommand{\agem}{\textsc{a-gem}\xspace}

\newcommand{\mer}{\textsc{mer}\xspace}
\newcommand{\er}{\textsc{er-reservoir}\xspace}
\newcommand{\errand}{\textsc{er-ringbuffer}\xspace}
\newcommand{\erkmeans}{\textsc{er-k-means}\xspace}
\newcommand{\erokmeans}{\textsc{er-k-means}\xspace}
\newcommand{\erherding}{\textsc{er-mof}\xspace}
\newcommand{\eroherding}{\textsc{er-mof}\xspace}
\newcommand{\ermixup}{\textsc{er-mixup}\xspace}

\usepackage[acronym,smallcaps,nowarn,section,nogroupskip,nonumberlist]{glossaries}
\glsdisablehyper{}
\newacronym{IL}{il}{incremental learning}
\newacronym{EWC}{ewc}{elastic weight consolidation}
\newacronym{PI}{pi}{path integral}
\newacronym{EWC++}{ewc++}{fast and online version of EWC}

\DeclareMathOperator*{\argmin}{argmin}

\algnewcommand{\LeftComment}[1]{\Statex #1}

\title{On Tiny Episodic Memories in Continual Learning}

\author{%
Arslan Chaudhry \\
\texttt{arslan.chaudhry@eng.ox.ac.uk} \\
University of Oxford \\
\And 
Marcus Rohrbach \\
Facebook AI Research \\
\And
Mohamed Elhoseiny \\ 
King Abdullah University of Science and Technology \\
\And 
Thalaiyasingam Ajanthan\\
Australian National University \\
\And 
Puneet K. Dokania  \\
University of Oxford \\
\And 
Philip H. S. Torr \\
University of Oxford \\
\And
Marc'Aurelio Ranzato \\
Facebook AI Research \\
}

\begin{document}

\maketitle

\begin{abstract}
In continual learning (CL), an agent learns from a stream of tasks leveraging prior experience to transfer knowledge to future tasks. It is an ideal framework to decrease the amount of supervision in the existing learning algorithms. But for a successful knowledge transfer, the learner needs to remember how to perform previous tasks. One way to endow the learner the ability to perform tasks seen in the past is to store a small memory, dubbed episodic memory, that stores few examples from previous tasks and then to replay these examples when training for future tasks. In this work, we empirically analyze the effectiveness of a very small episodic memory in a CL setup where each training example is only seen once. Surprisingly, across four rather different supervised learning benchmarks adapted to CL, a very simple baseline, that jointly trains on both examples from the current task as well as examples stored in the episodic memory, significantly outperforms specifically designed CL approaches with and without episodic memory.
Interestingly, we find that repetitive training on even tiny memories of past tasks does not harm generalization, on the contrary, it improves it, with gains between 7\% and 17\% when the memory is populated with a single example per class.\footnote{Code: \url{https://github.com/facebookresearch/agem}} 
 \end{abstract}

\section{Introduction} \label{sec:intro}
The objective of continual learning (CL) is to rapidly learn new skills from a sequence of tasks leveraging the knowledge accumulated in the past. 
Catastrophic forgetting~\citep{mccloskey1989catastrophic}, i.e. the inability of a model to recall how to perform tasks seen in the past, makes such efficient 
adaptation extremely difficult.

This decades old problem of CL~\citep{ring1997child, thrun1998lifelong} is now seeing a surge of interest in the research community with several methods proposed to tackle catastrophic forgetting \citep{Rebuffi16icarl,Kirkpatrick2016EWC,Zenke2017Continual,lee2017lifelong,aljundi2017memory,lopez2017gradient,lee2017IMM,chaudhry2019agem}. In this work, we quantitatively study some of these methods (that assume a fixed network architecture) on four benchmark datasets under the following assumptions: i) each task is fully supervised, ii) each example from a task can only be seen once using the learning protocol proposed by~\citet{chaudhry2019agem} (see \textsection\ref{sec:protocol}), and iii) 
the model has access to a small memory storing examples of past tasks. Restricting the size of such episodic memory is important because it makes the continual learning problem more realistic and distinct from multi-task learning where complete datasets of all the tasks are available at each step.  

We empirically observe that a very simple baseline, dubbed Experience Replay (ER)\footnote{For consistency to prior work in the literature, we will refer to this approach which trains on the episodic memory as ER, although its usage for supervised learning tasks is far less established.}, that jointly trains on both the examples from the current task and examples stored in the very small episodic memory not only gives superior performance over the existing state-of-the-art approaches specifically designed for CL (with and without episodic memory), but it also is computationally very efficient. We verify this finding on four rather different supervised learning benchmarks adapted for CL; Permuted MNIST, Split CIFAR, Split miniImageNet and Split CUB. Importantly, repetitive training on the same examples of a tiny episodic memory does not harm generalization on past tasks. In \textsection\ref{sec:analysis}, we analyze this phenomenon and provide insights as to why directly training on the episodic memory does not have a detrimental effect in terms of generalization. Briefly, we observe that the training on the datasets of subsequent tasks acts like a data-dependent regularizer on past tasks allowing the repetitive training on tiny memory to generalize beyond the episodic memory. We further observe that methods, that do not train directly on the memory, such as GEM~\citep{lopez2017gradient} and A-GEM~\citep{chaudhry2019agem}, underfit the training data and end up not fully utilizing the beneficial effects of this implicit and data depdendent regularization.



Overall, ER with tiny episodic memories offers very strong performance at a very small additional computational cost over the  fine-tuning baseline. 
We believe that this approach will serve as a stronger baseline for the development of future CL approaches. 

\section{Related Work} \label{sec:rel_work}
\paragraph{Regularization-based CL approaches} These works attempt to reduce forgetting by regularizing the objective such that it either penalizes the feature drift on already learned tasks \citep{li2016learning,Rebuffi16icarl} or discourages change in parameters that were important to solve past tasks \citep{Kirkpatrick2016EWC,Zenke2017Continual,chaudhry2018riemannian,aljundi2017memory}. The former approach relies on the storage of network activations and subsequent deployment of knowledge distillation \citep{hinton2015distilling}, whereas the latter approach stores a measure of parameter importance whose best case memory complexity is the same as the total number of network parameters.

\paragraph{Memory-Based CL approaches} These approaches \citep{lopez2017gradient,riemer2018learning,chaudhry2019agem} use episodic memory that stores a subset of data from past tasks to tackle forgetting. One approach to leverage such episodic memory is to use it to constrain the optimization such that the loss on past tasks can never increase~\citep{lopez2017gradient}. 

\paragraph{Experience Replay (ER)} The use of ER is well established in reinforcement learning (RL) tasks \citep{mnih2013playing,mnih2015human,foerster2017stabilising, rolnick18}. \citet{isele2018selective}, for instance, explore different ways to populate a relatively large episodic memory for a continual RL setting where the learner does multiple passes over the data. In this work instead, we study supervised learning tasks with a single pass through data and a very small episodic memory. More recently, \citep{hayes2018memory,riemer2018learning} used ER for supervised CL learning tasks. \citet{hayes2018memory}, independently, study different replay strategies in ER and show improvements over the finetune baseline. Our contribution is to show the improvements brought by ER, perhaps surprisingly, over the specifically designed CL approaches. We differ from~\citep{riemer2018learning} in considering episodic memories of much smaller sizes. Finally, and most importantly, we extend these previous studies by analyzing {\em why} repetitive training on tiny memories does not lead to overfitting (\textsection\ref{sec:analysis}).

\section{Learning Framework} \label{sec:protocol}

\subsection{Protocol for Single-Pass Through the Data} \label{subsec:proto}

We use the learning protocol proposed by~\citet{chaudhry2019agem}. There are two streams of tasks, described by the following ordered sequences of datasets, one stream for {\em C}ross-{\em V}alidation $\dataset^{CV} = \{\dataset_{-T^{CV}}, \cdots,\dataset_{-1}\}$ consisting of $T^{CV}$ tasks, and one for {\em EV}aluation $\dataset^{EV} = \{\dataset_{1}, \cdots,\dataset_T\}$ consisting of $T$ tasks, where $\dataset_k = \{(\ip_i^k, t_i^k, y_i^k)_{i=1}^{n_k}\}$ is the dataset of the $k$-th task. The sequence $\dataset^{CV}$ contains only a handful of tasks and it is only used for cross-validation purposes. Tasks from this sequence can be replayed as many times as needed and have various degree of similarity to tasks in the training and evaluation dataset, $\dataset^{EV}$. The latter stream, $\dataset^{EV}$, instead can be played only once; the learner will observe examples in {\em sequence} and will be tested throughout the learning experience. The final performance will be reported on the held-out test set drawn from  $\dataset^{EV}$.

The $k$-th task in any of these streams consists of $\dataset_k = \{(\ip_i^k, t_i^k, y_i^k)_{i=1}^{n_k}\}$, where each triplet 
constitutes an example defined by an input ($\ip^k \in \ipspace$), a task descriptor ($t^k \in \descspace$) which is an integer id in this work, and a target vector ($y^k \in \op^k$), where $\op^k$ is the set of labels specific to task $k$ and $\op^k \subset \opspace$.

\subsection{Metrics} \label{sec:metrics}
We measure performance on $\dataset^{EV}$ using two metrics, as standard practice in the literature~\citep{lopez2017gradient,chaudhry2018riemannian}:

\paragraph{Average Accuracy ($A \in [0,1]$)} Let $a_{i, j}$ be the performance of the model on the held-out test set of task `$j$' after the model is trained on task `$i$'. The average accuracy at task $T$ is then defined as:
\small
\begin{equation} \label{eq: avg_acc}
    A_{T} = \frac{1}{T} \sum_{j=1}^T a_{T, j}
\end{equation}
\normalsize
\paragraph{Forgetting ($F \in [-1,1]$)} Let $f_j^i$ be the forgetting on task `$j$' after the model is trained on task `$i$' which is computed as:
\small
\begin{equation}
        f_j^i = \max_{l \in \{1,\cdots,i-1\}} a_{l, j} - a_{i, j}
\end{equation}
\normalsize
The average forgetting measure at task $T$ is then defined as:
\small
\begin{equation} \label{eq:fgt}
        F_T = \frac{1}{T-1} \sum_{j=1}^{T-1} f_j^T
\end{equation}
\normalsize
\vspace{-3mm}
\section{Experience Replay} \label{sec:method}

Recent works~\citep{lopez2017gradient, chaudhry2019agem} have shown that methods relying on episodic memory have superior performance than regularization based 
approaches (e.g., ~\citep{Kirkpatrick2016EWC,Zenke2017Continual}) when using a ``single-pass through the data'' protocol (\textsection\ref{subsec:proto}). 
While GEM~\citep{lopez2017gradient} and its more efficient version A-GEM~\citep{chaudhry2019agem} used the episodic memory as a mean to project gradients, here we drastically simplify the optimization problem and, similar to \citet{riemer2018learning} and \citet{hayes2018memory}, directly train on the the examples stored in a very small memory, resulting in better performance and more efficient learning. 


The overall training procedure is given in Alg.~\ref{alg:main_er}. Compared to the simplest baseline model that merely fine-tunes the parameters on the new task starting from the 
previous task parameter vector, ER makes two modifications. First, it has an episodic memory which is updated at every time step, \texttt{line 8}. 
Second, it doubles the size of the minibatch used to compute the gradient descent parameter update  by stacking the actual minibatch of examples from the current task
with a minibatch of examples taken at random from the memory, \texttt{line 7}.
As we shall see in our empirical validation, these two simple modifications yield much better generalization and substantially limit forgetting, while incurring in
a negligible additional computational cost on modern GPU devices. Next, we explain the difference between the direct (ER) and indirect (A-GEM) training on episodic memory from the optimization perspective. 

\paragraph{A-GEM vs ER:} Let us assume that $B_n$ is a mini-batch of size $K$ from the current task $t$ and $B_{\epsmem}$ is the same size mini-batch from a very small episodic memory $\epsmem$. Furthermore, following the notation from~\cite{chaudhry2019agem}, let $g$ be the gradient computed with mini-batch $B_n$ and $g_{ref}$ be the gradient computed with $B_{\epsmem}$. In A-GEM, if $g^T g_{ref}\ge 0$, then the current task gradient $g$ is directly used for optimization whereas if $g^T g_{ref}< 0$, $g$ is projected such that $g^T g_{ref}= 0$. Refer to Eq.~11 in~\cite{chaudhry2019agem} for the exact form of projection. In ER instead, since both  mini-batches are used in the optimization step, the {\em average} of $g$ and $g_{ref}$ is used.  
It may seem a bit counter-intuitive that, even though ER repetitively trains on $\epsmem$, it is still able to generalize to previous tasks beyond the episodic memory. We investigate this question in \textsection\ref{sec:analysis}. 

Since we study the usage of tiny episodic memories, the sample that the learner selects to populate the memory becomes crucial, see \texttt{line 8} of the algorithm. For this, we describe various strategies to write into the memory. 
All these strategies assume access to a {\em continuous} stream of data and a small episodic memory, which rules out approaches relying on the temporary storage of all the examples seen so far. This restriction is consistent with our definition of CL: a learning experience through a stream of data under the constraint of a fixed and small sized memory and limited compute budget.

\begin{algorithm}[!t]
\caption{{\bf Experience Replay for Continual Learning.}}
\begin{algorithmic}[1]
  \Procedure{ER}{$\dataset, \mbox{mem\_sz}, \mbox{batch\_sz}, \mbox{lr}$}
    \State $\epsmem\gets \{\}*\mbox{mem\_sz}$ \Comment{{\tiny Allocate memory buffer of size mem\_sz}}
    \State $n \gets 0$ \Comment{{\tiny Number of training examples seen in the continuum}}
    \For{$t \in \{1, \cdots, T\}$}
        \For{$B_n \stackrel{K}{\sim} \dataset_t$} \Comment{{\tiny Sample without replacement a mini-batch of size $K$ from task $t$}}
        \State $B_{\epsmem} \stackrel{K}{\sim} \epsmem$ \Comment{{\tiny Sample a mini-batch from $\epsmem$}}
        \State $\param \gets SGD({B_n \cup B_{\epsmem}}, \param, \mbox{lr})$ \Comment{{\tiny Single gradient step to update the parameters by stacking current minibatch with minibatch from memory}}
        \State $\epsmem \gets \texttt{UpdateMemory}(\mbox{mem\_sz}, t, n, B_n)$ \Comment{{\tiny Memory update, see \textsection\ref{sec:method}}}
        \State $n \gets n+\mbox{batch\_sz}$ \Comment{{\tiny Counter update}}
        \EndFor
    \EndFor
    \State \textbf{return} $\param, \epsmem$
  \EndProcedure
\end{algorithmic}
\label{alg:main_er}
\end{algorithm}

\paragraph{Reservoir Sampling:} Similarly to~\citet{riemer2018learning}, Reservoir sampling~\citep{vitter1985random} takes as input a stream of data of unknown length 
and returns a random subset of items from that stream.
If `$n$' is the number of points observed so far and `\mbox{mem\_sz}' is the size of the reservoir (sampling buffer), this selection strategy samples each data point with a probability $\frac{\mbox{mem\_sz}}{n}$. The routine to update the memory is given in Appendix Alg.~\ref{alg:reservoir}.

\paragraph{Ring Buffer:} Similarly to~\citet{lopez2017gradient}, for each task, the ring buffer strategy allocates as many equally sized FIFO buffers as there are classes. 
If $C$ is the total number of classes across all tasks, and \mbox{mem\_sz} is the total size of episodic memory, each stack has a buffer of size $\frac{\mbox{mem\_sz}}{C}$.
As shown in Appendix Alg.~\ref{alg:ring}, the memory stores the last few observations from each class. Unlike reservoir sampling, samples from older tasks do not change 
throughout training, leading to potentially stronger overfitting. Also, at early stages of training the memory is not fully utilized since each stack has a constant size 
throughout training.
However, this simple sampling strategy guarantees equal representation of all classes in the 
memory, which is particularly important when the memory is tiny.

\paragraph{k-Means:} For each class, we use online k-Means to estimate the k centroids in feature space, using the representation before the last classification layer. 
We then store in the memory the input examples whose feature representation is the closest to such centroids, see Appendix Alg.~\ref{alg:kmeans}. This memory writing strategy
has similar benefits and drawbacks of ring buffer, except that it has potentially better coverage of the feature space in L2 sense.  

\paragraph{Mean of Features (MoF):} Similarly to \citet{Rebuffi16icarl}, for each class we compute a running estimate of the average feature vector 
just before the classification layer and store examples whose feature representation is closest to the average feature vector (see details in Appendix Alg.~\ref{alg:mof}). This writing strategy has the same balancing guarantees of ring buffer and k-means, but it populates the memory differently.
Instead of populating the memory at random or using k-Means, it puts examples that are closest to the mode in feature space.


\section{Experiments} \label{sec:experiments}
In this section, we review the benchmark datasets used in our evaluation, as well as the architectures and the baselines we compared against.
We then report the results we obtained using episodic memory and experience replay (ER). Finally, we conclude with a brief analysis investigating generalization when using 
ER on tiny memories.

\subsection{Datasets}
We consider four commonly used benchmarks in CL literature. \textbf{Permuted MNIST}~\citep{Kirkpatrick2016EWC} is a variant of MNIST~\citep{lecun1998mnist} 
dataset of handwritten digits where each task has a certain random permutation of the input pixels which is applied to all the images of that task. 
Our Permuted MNIST benchmark consists of a total of $23$ tasks. 

\textbf{Split CIFAR}~\citep{Zenke2017Continual} consists of splitting the original CIFAR-100 dataset~\citep{krizhevsky2009learning} into $20$ disjoint subsets, 
each of which is considered as a separate task. Each task has $5$ classes that are randomly sampled {\em without} replacement from the total of $100$ classes. 

Similarly to Split CIFAR, \textbf{Split miniImageNet} is constructed by splitting miniImageNet~\citep{vinyals2016matching}, a 
subset of ImageNet with a total of 100 classes and 600 images per class, to $20$ disjoint subsets. 

Finally, \textbf{Split CUB}~\citep{chaudhry2019agem} is an incremental version of the fine-grained image classification dataset CUB~\citep{WahCUB_200_2011} 
of $200$ bird categories split into $20$ disjoint subsets of classes. 

In all cases,  $\dataset^{CV}$ consists of $3$ tasks while $\dataset^{EV}$ contains the remaining tasks. As described in \textsection~\ref{sec:metrics}, we report metrics on 
$\dataset^{EV}$ after doing a single training pass over each task in the sequence. The hyper-parameters selected via cross-validation on $\dataset^{CV}$ are reported 
in Appenddix Tab.~\ref{tab:hyper_params}.

\subsection{Architectures}
For MNIST, we use a fully-connected network with two hidden layers of 256 ReLU units each. 
For CIFAR and miniImageNet, a reduced ResNet18, similar to~\citet{lopez2017gradient}, is used and a standard ResNet18 with ImageNet pretraining is used for CUB. 
The input integer task id is used to select a task specific classifier head, and the network is trained via cross-entropy loss.

For a given dataset stream, all baselines use the same architecture,
and all baselines are optimized via stochastic gradient descent with a mini-batch size equal to 10. 
The size of the mini-batch sampled from the episodic memory is also set to 10 irrespective of the size of the episodic buffer.

\subsection{Baselines}
We compare against the following baselines:
\begin{itemize}
\item \van, a model trained continually without any regularization and episodic memory, with parameters of a  new task initialized 
from the parameters of the previous task. 
    \item \ewc~\citep{Kirkpatrick2016EWC}, a regularization-based approach that avoids catastrophic forgetting by limiting the learning of parameters 
critical to the performance of past tasks, as measured by the Fisher information matrix (FIM). In particular, we compute the FIM as a moving average similar to \ewcp in~\cite{chaudhry2018riemannian} and online EWC in~\cite{progresscompress}.    
    \item \agem~\citep{chaudhry2019agem}, a model that uses episodic memory as an optimization constraint to avoid catastrophic forgetting. Since \gem \citep{lopez2017gradient} and \agem have similar performance, we only consider the latter in our experiments due to its computational efficiency.
    \item \mer~\citep{riemer2018learning}, a model that also leverages an episodic memory and uses a loss that approximates the dot products of the gradients of current and previous tasks to avoid forgetting. To make the experimental setting more comparable (in terms of SGD updates) to the other methods, we set the number of inner gradient steps to $1$ for each outer Reptile~\citep{metareptile} meta-update with the mini-batch size of $10$.
\end{itemize}

\subsection{Results} \label{sec:results}

\begin{figure}[!th]
    \begin{center}
    \begin{subfigure}{0.45\linewidth}
        \begin{center}
                \includegraphics[scale=0.42]{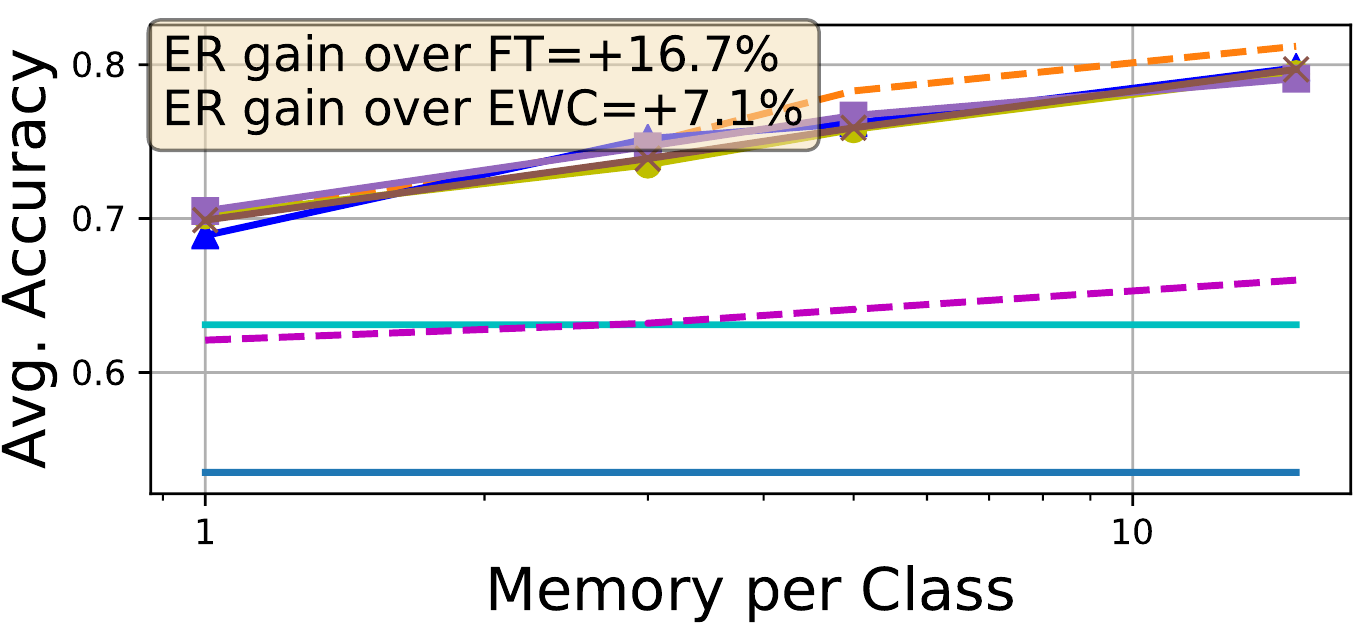}
                \caption{\small MNIST}
        \end{center}
        \end{subfigure}                                                                                                                                     
        \begin{subfigure}{0.45\linewidth}
        \begin{center}
                \includegraphics[scale=0.42]{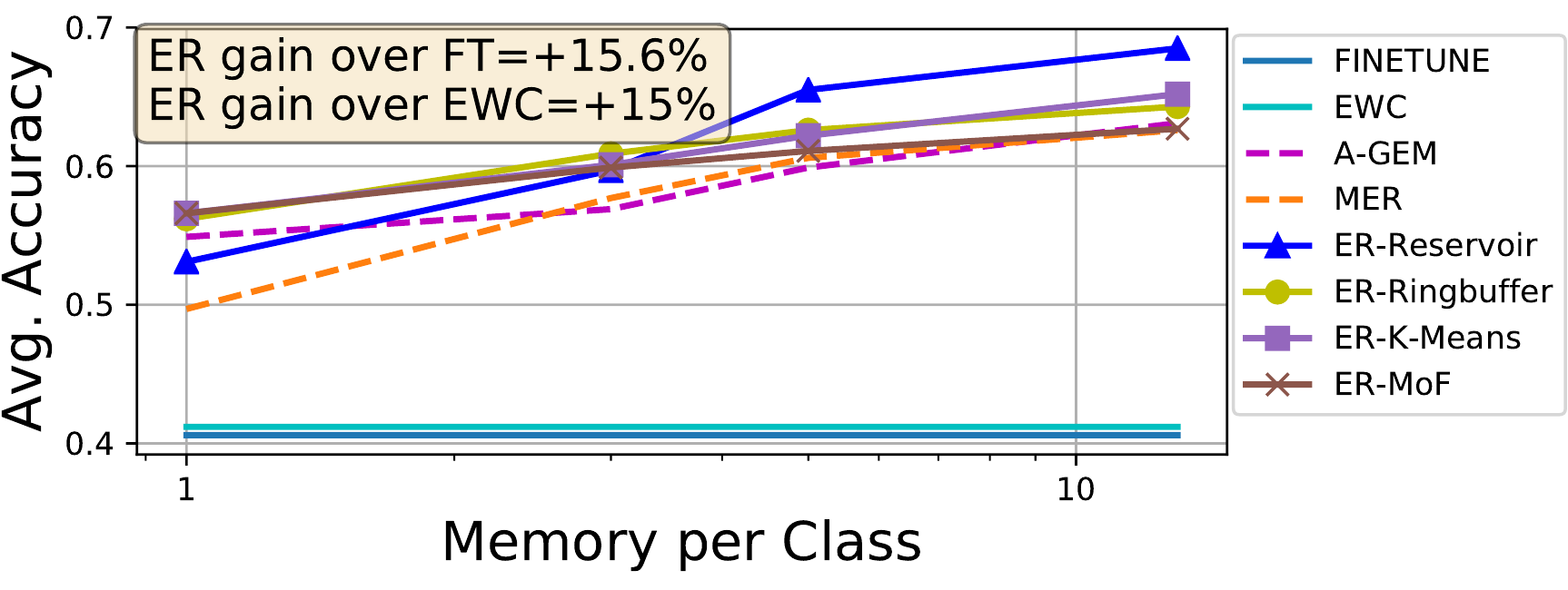}
                \caption{\small CIFAR}
                \end{center}
        \end{subfigure}
        \begin{subfigure}{0.45\linewidth}
        \begin{center}
                \includegraphics[scale=0.42]{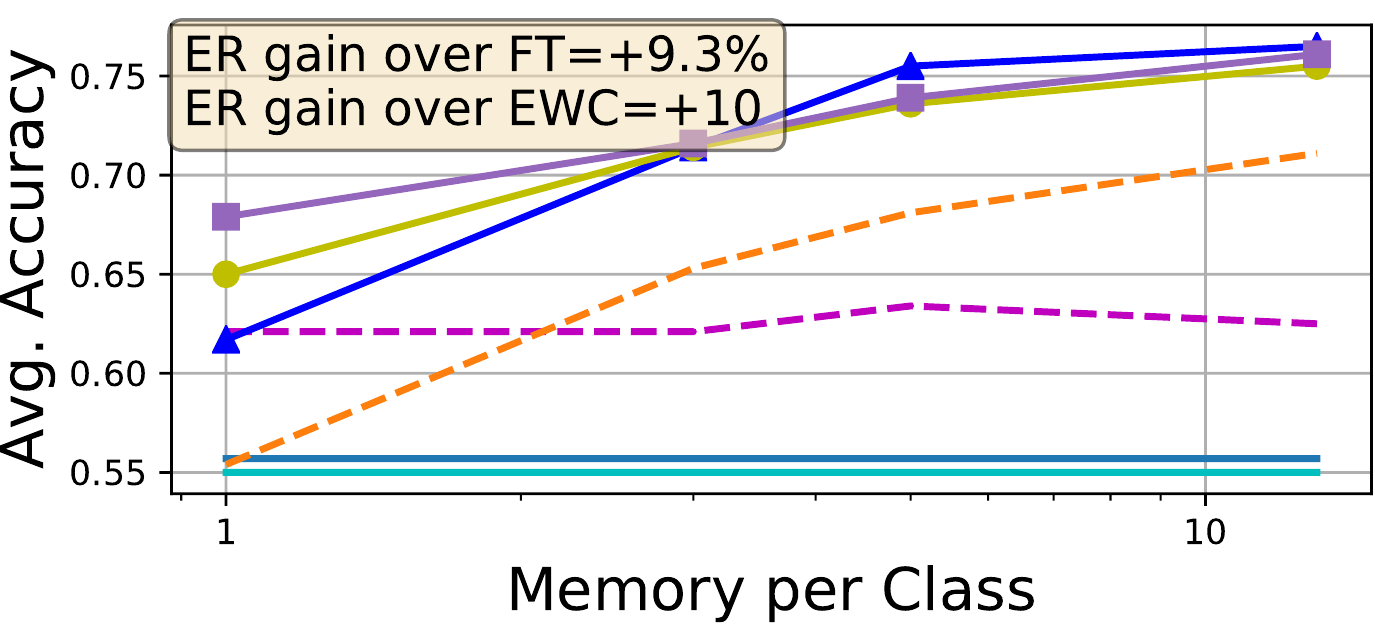}
                \caption{\small CUB}
        \end{center}
        \end{subfigure}
        \begin{subfigure}{0.45\linewidth}
        \begin{center}
                \includegraphics[scale=0.42]{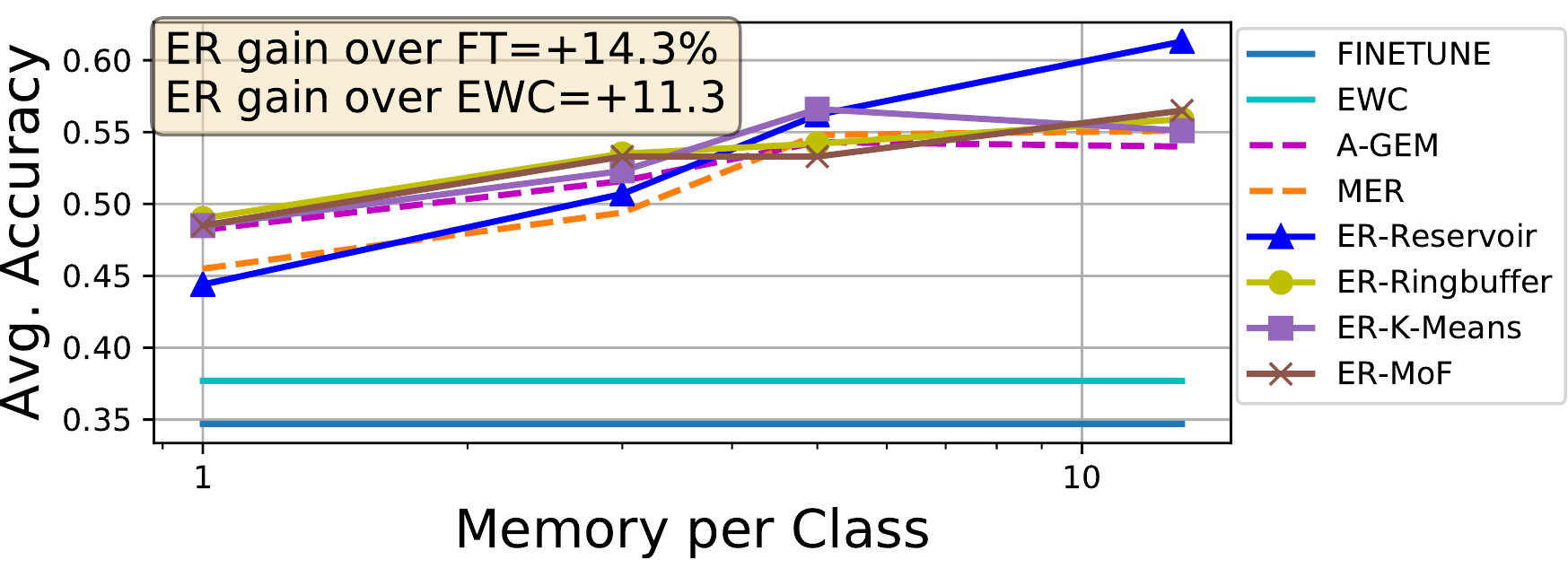}
                \caption{\small miniImageNet}
        \end{center}
        \end{subfigure}
         \end{center}
        \vspace{-3mm}
\caption{\em \small Average accuracy as a function of episodic memory size. The box shows the gain in average accuracy of {\errand} over {\van} and {\ewc} baselines when only $1$ sample per class is used. The performance is averaged over $5$ runs. Uncertainty estimates are provided in Appendix Tabs~\ref{tab:main_mnist_comp},\ref{tab:main_cifar_comp},\ref{tab:main_imagenet_comp},\ref{tab:main_cub_comp}.}
        \label{fig:all_increasing_samples}
\end{figure}

\begin{figure}[t]
    \begin{center}
                \includegraphics[scale=0.40]{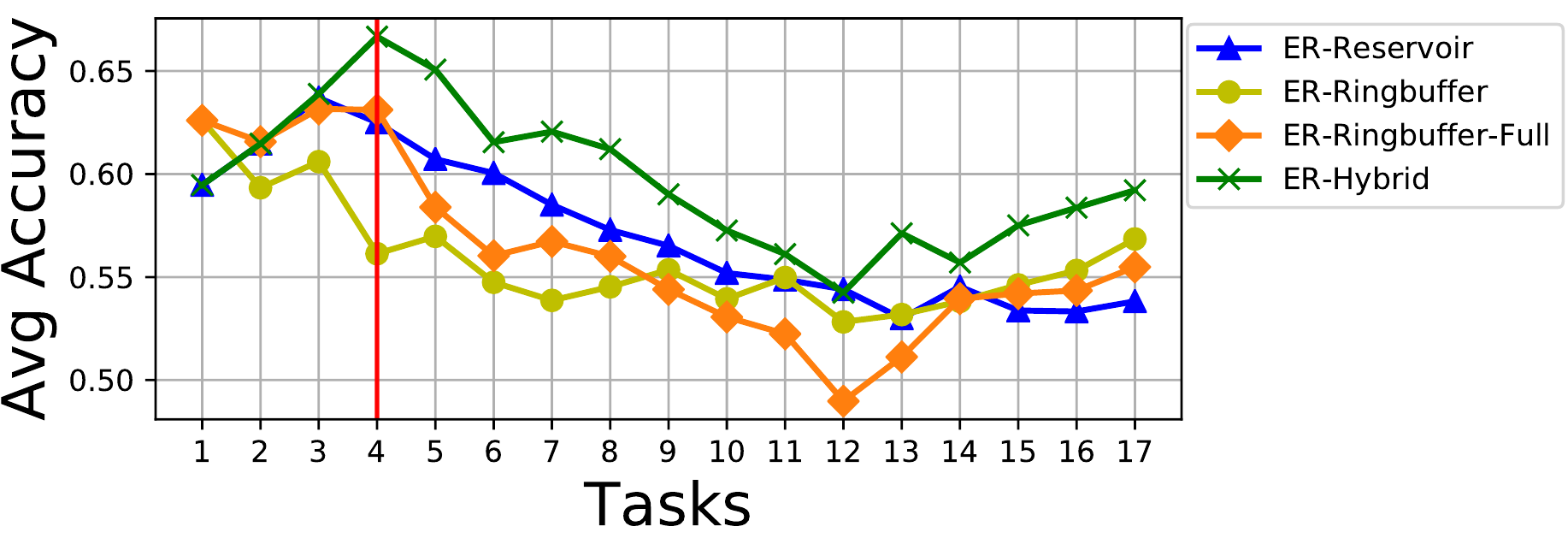}
        \end{center}
        \vspace{-3mm}
\caption{\small \em Evolution of average accuracy ($A_k$) as new tasks are learned in Split CIFAR. The memory has only $85$ slots (in average 1 slot per class).  The vertical bar marks 
where the hybrid approach switches from reservoir to ring buffer strategy. The hybrid approach works better than both reservoir (once more tasks arrive) and 
ring buffer (initially, when the memory is otherwise not well utilized). The orange curve is a variant of ring buffer that utilizes the full memory at all times, by 
reducing the ring buffer size of observed classes as new classes arrive. 
Overall, the proposed hybrid approach works at least as good as the other approaches throughout the whole learning experience. (Averaged over $3$ runs).
}
        \label{fig:hybrid_average_accuracy}
\end{figure}

In the first experiment, we measured average accuracy at the end of the learning experience on $\dataset^{EV}$ as a function of the size of the memory (detailed numerical results are provided in Appendix Tabs~\ref{tab:main_mnist_comp},\ref{tab:main_cifar_comp},\ref{tab:main_imagenet_comp},\ref{tab:main_cub_comp}). 
From the results in Fig.~\ref{fig:all_increasing_samples}, we can make several observations. 

First, methods using {\em ER greatly outperform} not only the baseline approaches that do not have episodic memory (\van and \ewc) but also state-of-the-art approaches relying on episodic memory of the same size (\agem and \mer). Moreover, the ER variants outperform even when the episodic memory is very small.
For instance, on CIFAR the gain over \agem brought by ER is 1.7\% when the memory only stores 1 example per class, and more than 5\% when the memory stores 13 examples per class.
This finding might seem quite surprising as repetitive training on a very small episodic memory may potentially lead to overfitting on the examples stored in the memory. We will investigate this finding in more depth in \textsection\ref{sec:analysis}. In the same setting, the gain compared to methods that do not use memory (\van and \ewc) is 15\% and about 28\% when using a single example per class and 13 examples per class, respectively.

Second and not surprisingly, average accuracy increases with the memory size, and does not saturate at 13 examples per class which is our self-imposed limit.

Third, experience replay based on reservoir sampling works the best across the board except when the memory size is very small (less than 3 examples per class).
Empirically we observed that as more and more tasks arrive and the size of the memory per class shrinks, reservoir sampling often ends up evicting some of the earlier 
classes from the memory, thereby inducing higher forgetting. 

Fourth, when the memory is tiny, sampling methods that by construction guarantee a balanced number of samples per class, work the best (even better than reservoir sampling). 
All methods that have this property, ring buffer, k-Means and Mean of Features, have a rather similar performance which is substantially better than the reservoir sampling.
For instance, on CIFAR, with one example per class in the memory, ER with reservoir sampling is 3.5\% worse than ER K-Means, but ER K-Means, ER Ring Buffer and ER MoF are all within 0.5\% from each other 
(see Appendix Tab.~\ref{tab:main_cifar_comp} for numerical values).  
These findings are further confirmed by looking at the evolution of the average accuracy (Appendix Fig.~\ref{fig:cifar_average_accuracy} left) as new tasks arrive when the memory can store at most one example per class.

\setlength{\tabcolsep}{1.0pt}
\begin{table}[!th]
\centering
\makebox[0pt][c]{\parbox{1.0\textwidth}{%
\begin{minipage}[t]{0.55\textwidth}
\small
\begin{tabular}{lcccc}
\toprule
\textbf{Methods}  & \multicolumn{4}{c}{\textbf{Forgetting}} \\
\hline
& \textbf{MNIST} & \textbf{CIFAR} & \textbf{CUB} & \textbf{miniImageNet}\\
\cmidrule(r){2-5}

{\van}           & 0.29 & 0.27 & 0.13 & 0.26 \\
{\ewc}           & 0.18 & 0.27 & 0.14 & 0.21 \\
{\agem}          & 0.21 & 0.14 & 0.09 & 0.13 \\
{\mer}           & 0.14 & 0.19 & 0.10 & 0.15 \\
{\errand} (ours) & \textbf{0.12} & \textbf{0.13} & \textbf{0.03} & \textbf{0.12} \\
\bottomrule
\end{tabular}
\captionof{table}{\em \small \textbf{Forgetting when using a tiny episodic memory of single example per class.}}
\label{tab:fgt_main}
\end{minipage}
\hfill
\begin{minipage}[t]{0.4\textwidth}
\small
\begin{tabular}{lcc}
\toprule
\textbf{Methods}  & \multicolumn{2}{c}{\textbf{Training Time [s]}} \\
\hline
& \textbf{CIFAR} & \textbf{CUB} \\
\cmidrule(r){2-3}

{\van} & 87 & 194 \\
{\ewc} & 159 & 235 \\
{\agem}   & 230 & 510 \\
{\mer} & 755 & 277  \\
{\errand} (ours) & 116 & 255 \\
\bottomrule
\end{tabular}
\captionof{table}{\em \small \textbf{Learning Time on $\dataset^{EV}$ [s]}}
\label{tab:ten_epochs_analysis}
\end{minipage}%
}}
\end{table}
\setlength{\tabcolsep}{6pt}

The better performance of strategies like ring buffer for tiny episodic memories, and reservoir sampling for bigger episodic memories, suggests a {\em hybrid} 
approach, whereby the writing strategy relies on reservoir sampling till some classes have too few samples stored in the memory. At that point, the writing strategy switches 
to the ring buffer scheme which guarantees a minimum number of examples for each class. 
For instance, in the experiment of Fig.~\ref{fig:hybrid_average_accuracy} the memory budget consists of only $85$ memory slots, an average of $1$ sample per class by the end of the learning experience (as there are $17$ tasks and $5$ classes per task). The learner switches 
from reservoir sampling to ring buffer once it observes that any of the classes seen in the past has only one sample left in the memory. 
When the switch happens (marked by a red vertical line in the figure), the learner only keeps randomly picked $\min(n,\frac{|\epsmem|}{K})$ examples per class in the memory, 
where $n$ is the number of examples of class $c$ in the memory and $K$ are the total number of classes observed so far. 
The overwriting happens opportunistically, removing examples
from over-represented classes as new classes are observed. Fig.~\ref{fig:hybrid_average_accuracy} shows that when the number of tasks is small, the hybrid version enjoys the high accuracy of reservoir sampling. As more tasks arrive
 and the memory per task shrinks, the hybrid scheme achieves superior performance than reservoir (and at least similar to ring buffer). 

Finally, experience replay methods are not only outperforming all other approaches in terms of accuracy 
(and lower forgetting as reported in Tab.~\ref{tab:fgt_main}), but also in terms of compute time. Tab.~\ref{tab:ten_epochs_analysis} reports training time on both Split CIFAR and Split CUB, using ring buffer as a use case since
all other ER methods have the same computational complexity. We observe that ER adds only a slight overhead compared to the finetuning baseline, but it is much cheaper than 
stronger baseline methods like \agem and \mer.

\subsection{Analysis} \label{sec:analysis}

The strong performance of experience replay methods which directly learn using the examples stored in the small episodic memory may be surprising. In fact, \citet{lopez2017gradient}
discounted this repetitive training on the memory option by saying: 
``Obviously, minimizing the loss at the current example together with [the loss on the episodic memory] results in overfitting to the
examples stored in [the memory]''. How can the repeated training over the same very small handful of examples possibly generalize? 

To investigate this matter we conducted an additional experiment. For simplicity, 
we consider only two tasks, $T_1$ and $T_2$, 
and study the generalization performance on $T_1$ as we train on $T_2$. 
We denote by $\dataset_2$ the training set of $T_2$ and by $\epsmem_1$ the memory storing examples from $T_1$'s training set. 
Our hypothesis is that although direct training on the examples in $\epsmem_1$ (in addition to those coming from $\dataset_2$) 
does indeed lead to strong memorization of $\epsmem_1$ (as measured by nearly zero cross-entropy loss on $\epsmem_1$), such training is still overall beneficial 
in terms of generalization on the original task $T_1$ because the joint learning with the examples of the current task $T_2$ acts as a strong, albeit 
implicit and data-dependent, regularizer for $T_1$.

\begin{figure}[t]
    \begin{center}
        \begin{subfigure}{0.30\linewidth}
        \begin{center}
                \includegraphics[scale=0.30]{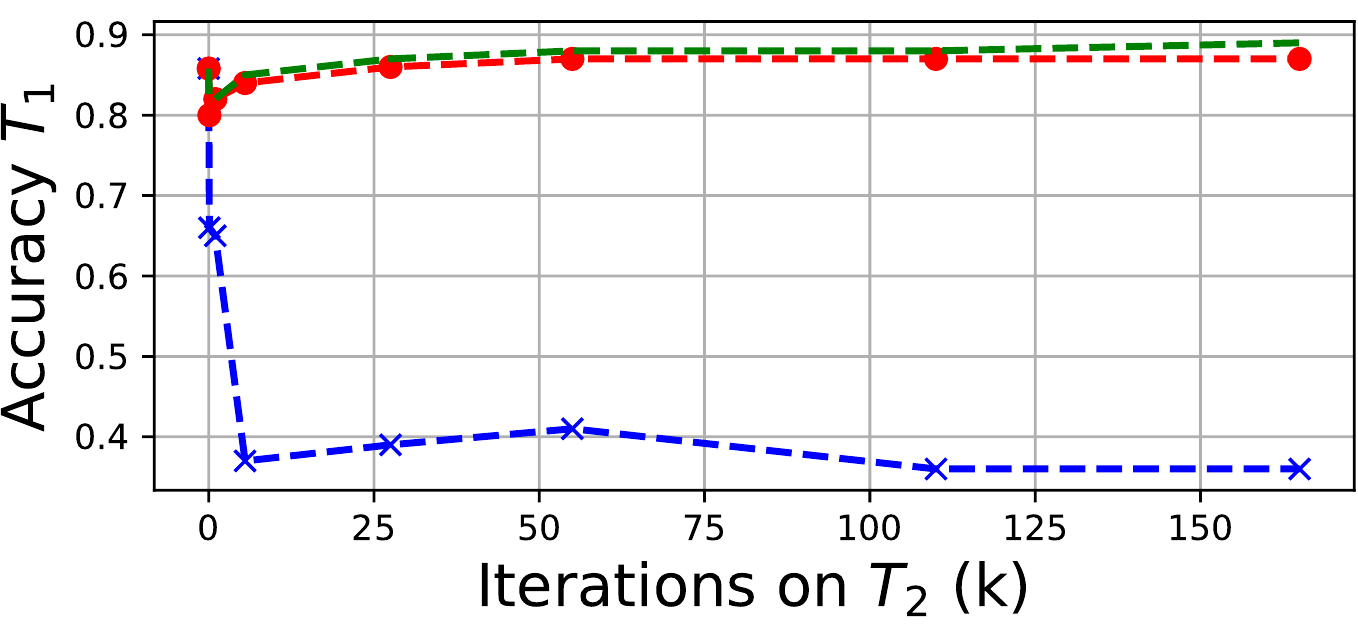}
                \caption{\small $20\degree$ rotation}
        \end{center}
        \end{subfigure}
        \begin{subfigure}{0.30\linewidth}
        \begin{center}
                \includegraphics[scale=0.30]{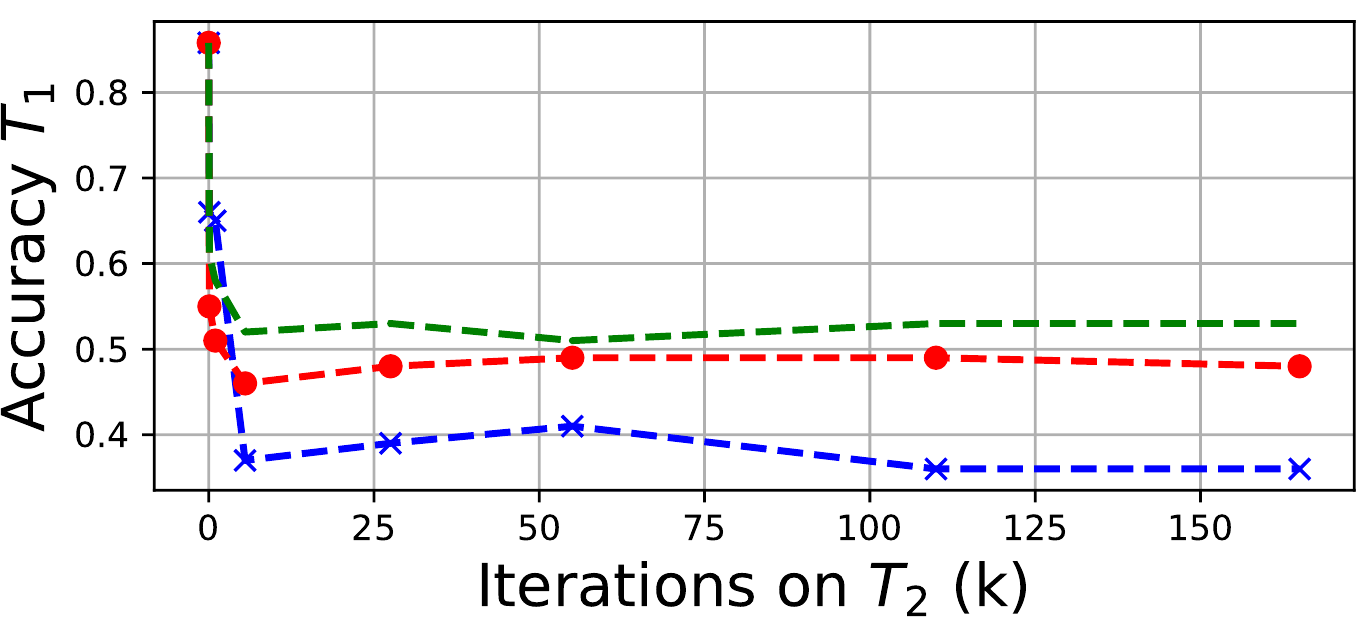}
                \caption{\small $40\degree$ rotation}
                \end{center}
        \end{subfigure}
        \begin{subfigure}{0.30\linewidth}
        \begin{center}
                \includegraphics[scale=0.30]{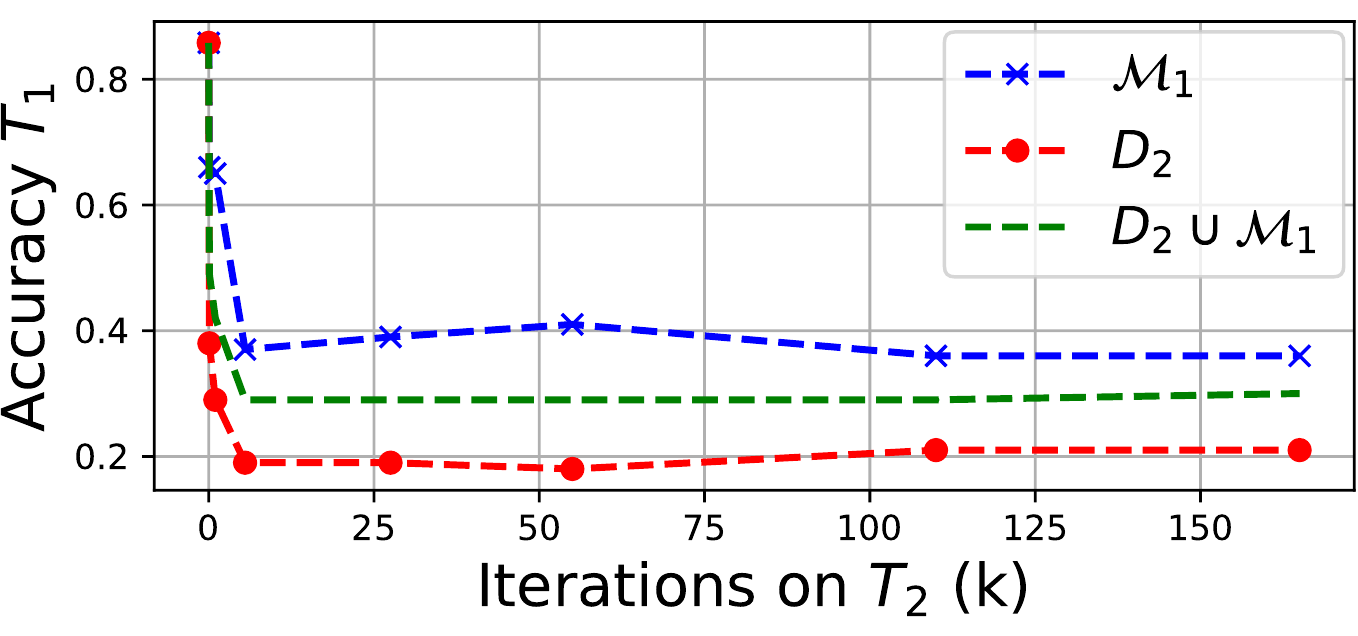}
                \caption{\small $60\degree$ rotation}
        \end{center}
        \end{subfigure}
    \end{center}
        \vspace{-3mm}
\caption{\em \small Analysis on MNIST Rotation: Test accuracy on Task 1 as a function of the training iterations over Task 2. The blue curves are the accuracy when the model is trained using only $\epsmem_1$. The red curves are the accuracy when the model is trained using only $\dataset_2$, the training set of Task 2. The green curves are the accuracy when in addition to $\dataset_2$, the model uses the memory from Task 1, $\epsmem_1$ (experience replay). (Averaged over $3$ runs).}
        \label{fig:mnist_rotations}
\end{figure}

To validate this hypothesis, we consider the MNIST Rotations dataset~\citep{lopez2017gradient}, where each task has digits rotated by a certain degree, a setting that enables fine control over the relatedness between the tasks. The architecture is the same as for Permuted MNIST, 
with only $10$ memory slots, one for each class of $T_1$. 
First, we verified that the loss on $\epsmem_1$ quickly drops to nearly $0$ as the model is trained using both $\epsmem_1$ and $\dataset_2$. 
As expected, the model achieves a perfect performance on the examples in the memory, which is not true for methods like \agem which make less direct use of the memory (see Appendix Tab.~\ref{tab:ten_mnist_rotation}).  
We then verified that only training on $\epsmem_1$ without $\dataset_2$, yields strong overfitting to the examples in the memory and poor generalization performance, 
with a mere average accuracy of $40\%$ on $T_1$ from the initial $85\%$ which was obtained just after training on $T_1$.
If we only train on $\dataset_2$ without using $\epsmem_1$ (same as \van baseline), we also observed overfitting to $\dataset_2$ as long as $T_2$ and $T_1$ are sufficiently 
unrelated, Fig.~\ref{fig:mnist_rotations}(b) and~\ref{fig:mnist_rotations}(c).   

When the two tasks are closely related instead (difference of rotation angles less than 20 degrees),  
we observe that even without the memory, generalization on $T_1$ improves 
as we train on $T_2$ because of positive transfer from the related task, see red curve in Fig.~\ref{fig:mnist_rotations}(a).
However, when we train on both $\dataset_2$ and $\epsmem_1$, generalization on $T_1$ is better than \van baseline, \ie, training with $\dataset_2$ only, regardless of the degree of relatedness between the two tasks, 
as shown by the green curves in Fig.~\ref{fig:mnist_rotations}. 

These findings suggest that while the model essentially memorizes the examples in the memory, this does not necessarily have a detrimental effect in terms of generalization 
as long as such learning is performed in conjunction with the examples of $T_2$. Moreover, there are two major axes controlling this regularizer: 
the number of examples in $T_2$ and
the relatedness between the tasks.
The former sets the strength of the regularizer. The latter, as measured by the accuracy on $T_1$
when training only on $\dataset_2$, controls its effectiveness. 
When $T_1$ and $T_2$ are closely related, Fig.~\ref{fig:mnist_rotations}(a), training on $\dataset_2$ prevents overfitting to $\epsmem_1$ by providing 
a data-dependent regularization that, even by itself, produces positive transfer. 
When $T_1$ and $T_2$ are somewhat related, Fig.~\ref{fig:mnist_rotations}(b), training on $\dataset_2$ still improves generalization on $T_1$ albeit to a much lesser extent.
However, when the tasks are almost adversarial to each other as an upside down 2 may look like a 5, 
the resulting regularization becomes even harmful, Fig.~\ref{fig:mnist_rotations}(c). In this case, accuracy drops from $40\%$ (training only on $\epsmem_1$) 
to $30\%$ (training on both $\epsmem_1$ and $\dataset_2$).

One remaining question related to generalization is how ER relates to A-GEM~\citep{chaudhry2019agem} and whether A-GEM overfits even less?
The answer is positive. As shown in Appendix Tab.~\ref{tab:ten_mnist_rotation}, A-GEM's accuracy on the memory examples does not reach 100\% even after having processed 1000 samples.
Interestingly, accuracy on the training set is lower than ER suggesting that the more constrained weight updates of A-GEM make it actually underfit. This underfitting prevents A-GEM from reaping the full regularization benefits brought by training on the data of subsequent tasks. 

\section*{Conclusions}
In this work we studied ER methods for supervised CL tasks. Our empirical analysis on several
benchmark streams of data shows that ER methods even with a tiny episodic memory offer a very large performance boost at a very marginal increase of computational cost compared to the finetuning baseline. We also studied various ways to populate the memory and proposed a hybrid approach that strikes a good trade-off between randomizing the examples in the memory while keeping enough representatives for each class. 

Our study also sheds light into a very interesting phenomenon: memorization (zero cross-entropy loss) of tiny memories is useful for generalization because training on subsequent tasks acts like a data dependent regularizer. 
Overall, we hope the CL community will adopt experience replay methods as a baseline, given their strong empirical performance, efficiency and simplicity of implementation.

There are several avenues of future work. For instance, we would like to investigate what are the optimal inputs that best mitigate expected forgetting and
optimal strategies to remove samples from the memory when it is entirely filled up.

\clearpage
\bibliographystyle{named}
\bibliography{chaudhryBibliography}

\clearpage
\section*{Appendix}
In \textsection{\ref{sec:mem_update}}, we provide algorithms for different memory update strategies described in \textsection{\ref{sec:method}} of the main paper. The detailed results of the experiments which were used to generate Fig.~\ref{fig:all_increasing_samples} and Tab.~\ref{tab:fgt_main} in the main paper are provided in \textsection{\ref{sec:detailed_results}}. The analysis conducted in \textsection{\ref{sec:analysis}} of the main paper is further described in \textsection{\ref{sec:further_analysis}}. Finally, in \textsection{\ref{sec:hyper_param}}, we list the hyper-parameters used for each of the baselines across all the datasets.  

\appendix

\section{Memory Update Algorithms} \label{sec:mem_update}

Here we provide the algorithms to write into memory as discussed in \textsection{\ref{sec:method}} of the main paper. 
\begin{algorithm}[!h]
\caption{{\bf Reservoir sampling update}. {\footnotesize mem\_sz is the number of examples the memory can store, $t$ is the task id, $n$ is the number of examples observed 
so far in the data stream, and $B$ is the input mini-batch.}}
\begin{algorithmic}[1]
  \Procedure{UpdateMemory}{$\mbox{mem\_sz}, t, n, B$}
  \State $j \gets 0$
  \For{$(\ip,y)$ in $B$}  
    \State $M \gets |\epsmem|$ \Comment{{\tiny Number of samples currently stored in the memory}}
    \If {$M < \mbox{mem\_sz}$}
        \State $\epsmem.\mbox{append}(\ip, y, t)$
    \Else
        \State $i = \mbox{randint}(0, n + j)$
        \If {$i < \mbox{mem\_sz}$}
            \State $\epsmem[i] \gets (\ip, y, t)$  \Comment{{\tiny Overwrite memory slot.}}
        \EndIf
    \EndIf
    \State $j \gets j + 1$
  \EndFor
  \State \textbf{return} $\epsmem$
  \EndProcedure
\end{algorithmic}
\label{alg:reservoir}
\end{algorithm}

\begin{algorithm}[!h]
\caption{{\bf Ring buffer}.}
\begin{algorithmic}[1]
  \Procedure{UpdateMemory}{$\mbox{mem\_sz}, t, n, B$}
    \For{$(\ip,y)$ in $B$}
    \State {\tiny \# Assume FIFO stacks $\epsmem[t][y]$ of fixed size are already initialized}
    \State $\epsmem[t][y].\mbox{append}(\ip)$
    \EndFor
  \State \textbf{return} $\epsmem$
  \EndProcedure
\end{algorithmic}
\label{alg:ring}
\end{algorithm}

\begin{algorithm}[!h]
\caption{{\bf K-Means}. {\footnotesize Memory is populated using samples closest (in feature space) to sequential K-Means centroids.}}
\begin{algorithmic}[1]
  \Procedure{UpdateMemory}{$\mbox{mem\_sz}, t, n, B$}
  \State {\tiny \# Assume array $\epsmem[t][y]$ of fixed size is already initialized}
  \State {\tiny \# Assume K centroids $c_j$ are already initialized}
  \State {\tiny \# Assume cluster counters $n_j$ are already initialized to $0$}
    \For{$(\ip,y)$ in $B$}
            \State $j \gets \argmin_{j \in \{1,\cdots,K\}} || \phi_{\param}(\ip) - c_j||$
            \State $n_j \gets n_j + 1$
            \State $c_j \gets c_j + \frac{1}{n_j}*(\phi_{\param}(\ip)-c_j)$
            \State $d = ||\phi_{\param}(\ip) - c_j||$
            \If{$d < \epsmem[t][y][j].\mbox{get\_dst()}$} \Comment{{\tiny Store the current example if it is closer to the centroid}}
                \State $\epsmem[t][y][j] \leftarrow (\ip, d)$
            \EndIf
    \EndFor
  \State \textbf{return} $\epsmem$
  \EndProcedure
\end{algorithmic}
\label{alg:kmeans}
\end{algorithm}

\begin{algorithm}[H]
\caption{{\bf Mean of Features}. {\footnotesize Store examples that are closest to the running average feature vector.}}
\begin{algorithmic}[1]
  \Procedure{UpdateMemory}{$\mbox{mem\_sz}, t, n, B$}
  \State {\tiny \# Assume heaps $\epsmem[t][y]$ of fixed size are already initialized}
  \State {\tiny \# Assume average features $f[t][y]$ are already initialized}
  \State {\tiny \# Assume moving average decay hyper-parameter ($\alpha$) is given}
    \For{$(\ip,y)$ in $B$}
      \State $f[t][y] \gets \alpha* f[t][y] + (1-\alpha)*\phi_{\param}(\ip)$
      \State $d = ||\phi_{\param}(\ip) - f[t][y]||$
      \If{ $\epsmem[t][y].\mbox{find\_max}() > d$} \Comment{{\tiny Store the current example if it is closer to the center}}
                \State $\epsmem[t][y].\mbox{delete\_max}()$
                \State $\epsmem[t][y].\mbox{insert}(\ip; d)$
      \EndIf
    \EndFor
  \State \textbf{return} $\epsmem$
  \EndProcedure
\end{algorithmic}
\label{alg:mof}
\end{algorithm}

\section{Detailed Results} \label{sec:detailed_results}

Here we describe the detailed results used to generate the Fig.~\ref{fig:all_increasing_samples} in the main paper. In addition we also report the forgetting metric~\eqref{eq:fgt}. Note that the {\mtask} baseline does not follow the definition of continual learning as it keeps the dataset of all the tasks around at every step. 

\begin{table}[!th]
\centering
\small
\caption{\em \textbf{Permuted MNIST}: Performance (average accuracy (left column) and forgetting (right column)) for different number of samples per class. The average accuracy numbers from the this table are used to generate Fig.~\ref{fig:all_increasing_samples} in \textsection\ref{sec:results} of the main paper.}
\label{tab:main_mnist_comp}
\resizebox{\textwidth}{!}{%
\begin{tabular}{lcccc|cccc}
\toprule
\multicolumn{1}{l}{\textbf{Methods}} &\multicolumn{8}{c}{\textbf{Episodic Memory (Samples Per Class)}} \\
\hline
        & \multicolumn{4}{c}{Average Accuracy [$A_{T}$(\%)]} & \multicolumn{4}{c}{Forgetting [$F_{T}$]} \\
        \cmidrule(r){2-5} \cmidrule(l){6-9}
            & 1 & 3 & 5 & 15 & 1 & 3 & 5 & 15 \\
        \cmidrule(r){2-5} \cmidrule(l){6-9}
{\agem}        & 62.1 (\textpm\ 1.39) & 63.2 (\textpm\ 1.47) & 64.1 (\textpm\ 0.74) & 66.0 (\textpm\ 1.78) & 0.21 (\textpm\ 0.01) & 0.20 (\textpm\ 0.01) & 0.19 (\textpm\ 0.01) & 0.17 (\textpm\ 0.02) \\
{\mer}         & 69.9 (\textpm\ 0.40) & 74.9 (\textpm\ 0.49) & 78.3 (\textpm\ 0.19) & 81.2 (\textpm\ 0.28) & 0.14 (\textpm\ 0.01) & 0.09 (\textpm\ 0.01) & 0.06 (\textpm\ 0.01) & 0.03 (\textpm\ 0.01) \\
{\errand}      & 70.2 (\textpm\ 0.56) & 73.5 (\textpm\ 0.43) & 75.8 (\textpm\ 0.24) & 79.4 (\textpm\ 0.43) & 0.12 (\textpm\ 0.01) & 0.09 (\textpm\ 0.01) & 0.07 (\textpm\ 0.01) & 0.04 (\textpm\ 0.01) \\
{\eroherding}  & 69.9 (\textpm\ 0.68) & 73.9 (\textpm\ 0.64) & 75.9 (\textpm\ 0.21) & 79.7 (\textpm\ 0.19) & 0.13 (\textpm\ 0.01) & 0.09 (\textpm\ 0.01) & 0.07 (\textpm\ 0.01) & 0.04 (\textpm\ 0.01) \\ 
{\erokmeans}   & 70.5 (\textpm\ 0.42) & 74.7 (\textpm\ 0.62) & 76.7 (\textpm\ 0.51) & 79.1 (\textpm\ 0.32) & 0.12 (\textpm\ 0.01) & 0.08 (\textpm\ 0.01) & 0.06 (\textpm\ 0.01) & 0.04 (\textpm\ 0.01) \\
{\er}          & 68.9 (\textpm\ 0.89) & 75.2 (\textpm\ 0.33) & 76.2 (\textpm\ 0.38) & 79.8 (\textpm\ 0.26) & 0.15 (\textpm\ 0.01) & 0.08 (\textpm\ 0.01) & 0.07 (\textpm\ 0.01) & 0.04 (\textpm\ 0.01) \\

\cmidrule(r){2-5} \cmidrule(l){6-9}
{\van} & 53.5 (\textpm\ 1.46) & - & - & - & 0.29 (\textpm\ 0.01) & - & - & \\
{\ewc} & 63.1 (\textpm\ 1.40) & - & - & - & 0.18 (\textpm\ 0.01) & - & - & \\
\cmidrule(r){2-5} \cmidrule(l){6-9}
{\mtask} & \multicolumn{4}{c}{83} & \multicolumn{4}{c}{-} \\
\bottomrule
\end{tabular}}           
\end{table}

\begin{table}[!th]
\centering
\small
\caption{\em \textbf{Split CIFAR}: Performance (average accuracy (left column) and forgetting (right column)) for different number of samples per class. The average accuracy numbers from the this table are used to generate Fig.~\ref{fig:all_increasing_samples} in \textsection\ref{sec:results} of the main paper.}
\label{tab:main_cifar_comp}
\resizebox{\textwidth}{!}{%
\begin{tabular}{lcccc|cccc}
\toprule
\multicolumn{1}{l}{\textbf{Methods}} &\multicolumn{8}{c}{\textbf{Episodic Memory (Samples Per Class)}} \\
\hline
        & \multicolumn{4}{c}{Average Accuracy [$A_{T}$(\%)]} & \multicolumn{4}{c}{Forgetting [$F_{T}$]} \\
        \cmidrule(r){2-5} \cmidrule(l){6-9}
            & 1 & 3 & 5 & 13 & 1 & 3 & 5 & 13 \\
        \cmidrule(r){2-5} \cmidrule(l){6-9}
{\agem}       & 54.9 (\textpm\ 2.92) & 56.9  (\textpm\ 3.45) & 59.9  (\textpm\ 2.64) & 63.1 (\textpm\ 1.24) & 0.14 (\textpm\ 0.03) & 0.13 (\textpm\ 0.03) & 0.10 (\textpm\ 0.02) & 0.07 (\textpm\ 0.01) \\ 
{\mer}   & 49.7 (\textpm\ 2.97) & 57.7 (\textpm\ 2.59) & 60.6 (\textpm\ 2.09) & 62.6 (\textpm\ 1.48) & 0.19 (\textpm\ 0.03) & 0.11 (\textpm\ 0.01) & 0.09 (\textpm\ 0.02) & 0.07 (\textpm\ 0.01) \\
{\errand} & 56.2 (\textpm\ 1.93) & 60.9 (\textpm\ 1.44) & 62.6 (\textpm\ 1.77) & 64.3 (\textpm\ 1.84) & 0.13 (\textpm\ 0.01) & 0.09 (\textpm\ 0.01) & 0.08 (\textpm\ 0.02) & 0.06 (\textpm\ 0.01) \\
{\eroherding}  & 56.6 (\textpm\ 2.09) & 59.9 (\textpm\ 1.25) & 61.1 (\textpm\ 1.62) & 62.7 (\textpm\ 0.63) & 0.12 (\textpm\ 0.01 ) & 0.10 (\textpm\ 0.01) & 0.08 (\textpm\ 0.01) & 0.07 (\textpm\ 0.01) \\
{\erokmeans} & 56.6 (\textpm\ 1.40) & 60.1 (\textpm\ 1.41) & 62.2 (\textpm\ 1.20) & 65.2 (\textpm\ 1.81) & 0.13 (\textpm\ 0.01) & 0.09 (\textpm\ 0.01) & 0.07 (\textpm\ 0.01) & 0.04 (\textpm\ 0.01) \\
{\er}  & 53.1 (\textpm\ 2.66) & 59.7 (\textpm\ 3.87)  & 65.5 (\textpm\ 1.99) & 68.5 (\textpm\ 0.65) & 0.19 (\textpm\ 0.02) & 0.12 (\textpm\ 0.03) & 0.09 (\textpm\ 0.02) & 0.05 (\textpm\ 0.01) \\
\cmidrule(r){2-5} \cmidrule(l){6-9}
{\van} & 40.6 (\textpm\ 3.83) & - & - & -  & 0.27 (\textpm\ 0.04)  & - & - & \\
{\ewc} & 41.2 (\textpm\ 2.67) & - & - & -  & 0.27 (\textpm\ 0.02)  & - & - & \\
\cmidrule(r){2-5} \cmidrule(l){6-9}
{\mtask} & \multicolumn{4}{c}{68.3} & \multicolumn{4}{c}{-} \\
\bottomrule
\end{tabular}}           
\end{table}

\begin{table}[!th]
\centering
\small
\caption{\em \textbf{miniImageNet}: Performance (average accuracy (left column) and forgetting (right column)) for different number of samples per class. The average accuracy numbers from the this table are used to generate Fig.~\ref{fig:all_increasing_samples} in \textsection\ref{sec:results} of the main paper.}
\label{tab:main_imagenet_comp}
\resizebox{\textwidth}{!}{%
\begin{tabular}{lcccc|cccc}
\toprule
\multicolumn{1}{l}{\textbf{Methods}} &\multicolumn{8}{c}{\textbf{Episodic Memory (Samples Per Class)}} \\
\hline
        & \multicolumn{4}{c}{Average Accuracy [$A_{T}$(\%)]} & \multicolumn{4}{c}{Forgetting [$F_{T}$]} \\
        \cmidrule(r){2-5} \cmidrule(l){6-9}
            & 1 & 3 & 5 & 13 & 1 & 3 & 5 & 13 \\
        \cmidrule(r){2-5} \cmidrule(l){6-9}
{\agem} &  48.2 (\textpm\ 2.49) & 51.6 (\textpm\ 2.69) & 54.3 (\textpm\ 1.56) & 54 (\textpm\ 3.63) & 0.13 (\textpm\ 0.02) & 0.10 (\textpm\ 0.02) & 0.08 (\textpm\ 0.01) & 0.09 (\textpm\ 0.03) \\ 
{\mer}  & 45.5 (\textpm\ 1.49) & 49.4 (\textpm\ 3.43) & 54.8 (\textpm\ 1.79) & 55.1 (\textpm\ 2.91) & 0.15 (\textpm\ 0.01) & 0.12 (\textpm\ 0.02) & 0.07 (\textpm\ 0.01) & 0.07 (\textpm\ 0.01) \\
{\errand} & 49.0 (\textpm\ 2.61) & 53.5 (\textpm\ 1.42)  & 54.2 (\textpm\ 3.23) & 55.9 (\textpm\ 4.05) & 0.12 (\textpm\ 0.02) & 0.07 (\textpm\ 0.02) & 0.08 (\textpm\ 0.02) & 0.06 (\textpm\ 0.03) \\
{\eroherding} & 48.5 (\textpm\ 1.72) & 53.3 (\textpm\ 2.80)  & 53.3 (\textpm\ 3.11) & 56.5 (\textpm\ 1.92) & 0.12 (\textpm\ 0.01) & 0.08 (\textpm\ 0.01) & 0.08 (\textpm\ 0.02) & 0.05 (\textpm\ 0.02) \\
{\erokmeans} & 48.5 (\textpm\ 0.35) & 52.3 (\textpm\ 3.12)  & 56.6 (\textpm\ 2.48) & 55.1 (\textpm\ 1.86) & 0.12 (\textpm\ 0.02) & 0.09 (\textpm\ 0.02) & 0.06 (\textpm\ 0.01) & 0.06 (\textpm\ 0.01) \\
{\er}  & 44.4 (\textpm\ 3.22) & 50.7 (\textpm\ 3.36) & 56.2 (\textpm\ 4.12) & 61.3 (\textpm\ 6.72) & 0.17 (\textpm\ 0.02) & 0.12 (\textpm\ 0.03) & 0.07 (\textpm\ 0.04) & 0.04 (\textpm\ 0.06) \\
\cmidrule(r){2-5} \cmidrule(l){6-9}
{\van} & 34.7 (\textpm\ 2.69) & -  & - & - & 0.26 (\textpm\ 0.03)  & - & - & \\
{\ewc} & 37.7 (\textpm\ 3.29) & -  & - & - & 0.21 (\textpm\ 0.03)  & - & - & \\
\cmidrule(r){2-5} \cmidrule(l){6-9}
{\mtask} & \multicolumn{4}{c}{62.4} & \multicolumn{4}{c}{-} \\
\bottomrule
\end{tabular}}           
\end{table}

\begin{table}[!th]
\centering
\small
\caption{\em \textbf{CUB}: Performance (average accuracy (left column) and forgetting (right column)) for different number of samples per class. The average accuracy numbers from the this table are used to generate Fig.~\ref{fig:all_increasing_samples} in \textsection\ref{sec:results} of the main paper.}
\label{tab:main_cub_comp}
\resizebox{\textwidth}{!}{%
\begin{tabular}{lcccc|cccc}
\toprule
\multicolumn{1}{l}{\textbf{Methods}} &\multicolumn{8}{c}{\textbf{Episodic Memory (Samples Per Class)}} \\
\hline
        & \multicolumn{4}{c}{Average Accuracy [$A_{T}$(\%)]} & \multicolumn{4}{c}{Forgetting [$F_{T}$]} \\
        \cmidrule(r){2-5} \cmidrule(l){6-9}
            & 1 & 3 & 5 & 10 & 1 & 3 & 5 & 10 \\
        \cmidrule(r){2-5} \cmidrule(l){6-9}
{\agem}        & 62.1 (\textpm\ 1.28) & 62.1 (\textpm\ 1.87) & 63.4 (\textpm\ 2.33) & 62.5 (\textpm\ 2.34) & 0.09 (\textpm\ 0.01) & 0.08 (\textpm\ 0.02) & 0.07 (\textpm\ 0.01) & 0.08 (\textpm\ 0.02) \\
{\mer}         & 55.4 (\textpm\ 1.03) & 65.3 (\textpm\ 1.68) & 68.1 (\textpm\ 1.61) & 71.1 (\textpm\ 0.93) & 0.10 (\textpm\ 0.01) & 0.04 (\textpm\ 0.01) & 0.03 (\textpm\ 0.01) & 0.03 (\textpm\ 0.01) \\
{\errand}      & 65.0 (\textpm\ 0.96) & 71.4 (\textpm\ 1.53) & 73.6 (\textpm\ 1.57) & 75.5 (\textpm\ 1.84) & 0.03 (\textpm\ 0.01) & 0.01 (\textpm\ 0.01) & 0.01 (\textpm\ 0.01) & 0.02 (\textpm\ 0.01) \\
{\erokmeans}   & 67.9 (\textpm\ 0.87) & 71.6 (\textpm\ 1.56) & 73.9 (\textpm\ 2.01) & 76.1 (\textpm\ 1.74) & 0.02 (\textpm\ 0.01) & 0.02 (\textpm\ 0.01) & 0.02 (\textpm\ 0.01) & 0.01 (\textpm\ 0.01) \\
{\er}          & 61.7 (\textpm\ 0.62) & 71.4 (\textpm\ 2.57) & 75.5 (\textpm\ 1.92) & 76.5 (\textpm\ 1.56) & 0.09 (\textpm\ 0.01) & 0.04 (\textpm\ 0.01) & 0.02 (\textpm\ 0.01) & 0.03 (\textpm\ 0.02) \\
\cmidrule(r){2-5} \cmidrule(l){6-9}
{\van} & 55.7 (\textpm\ 2.22) & - & - & - & 0.13 (\textpm\ 0.03) & - & - & \\
{\ewc} & 55.0 (\textpm\ 2.34) & - & - & - & 0.14 (\textpm\ 0.02) & - & - & \\
\cmidrule(r){2-5} \cmidrule(l){6-9}
{\mtask} & \multicolumn{4}{c}{65.6} & \multicolumn{4}{c}{-} \\
\bottomrule
\end{tabular}}           
\end{table}

\begin{figure}[!th]
	\begin{subfigure}{0.45\linewidth}
	\begin{center}
		\includegraphics[scale=0.42]{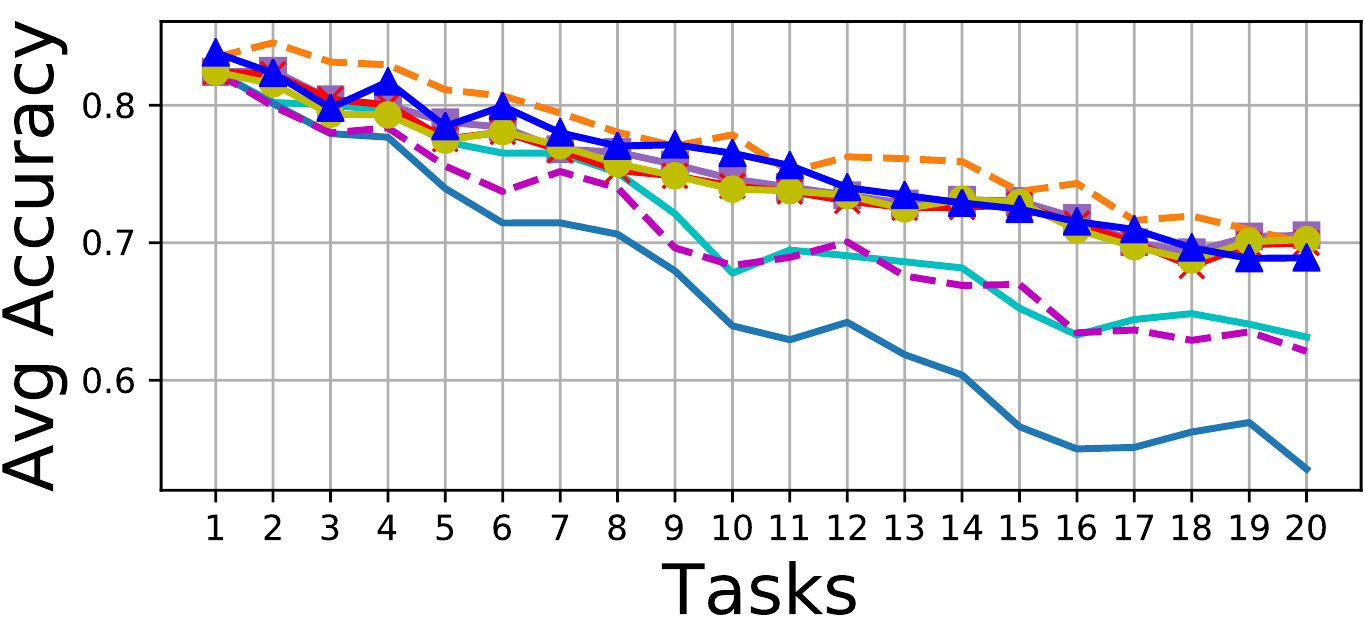}
		\caption{\small 1 Sample}
	\end{center}
	\end{subfigure}%
	\begin{subfigure}{0.45\linewidth}
	\begin{center}
		\includegraphics[scale=0.42]{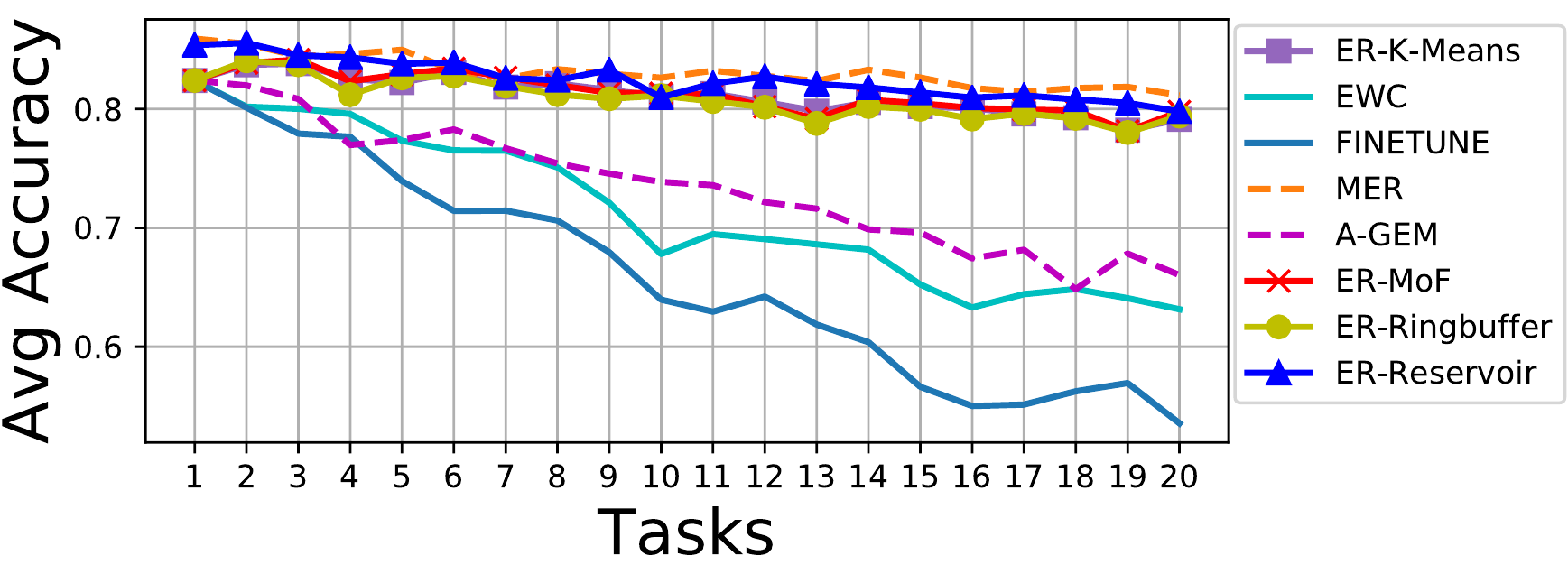}
		\caption{\small 15 Samples}
	\end{center}
	\end{subfigure}%
\caption{\em MNIST: Evolution of average accuracy ($A_k$) as new tasks are learned when `1' and `15' samples per class are used.}
	\label{fig:mnist_avg_accuracy}
\end{figure}

\begin{figure}[!th]
    \begin{center}
        \begin{subfigure}{0.45\linewidth}
        \begin{center}
                \includegraphics[scale=0.42]{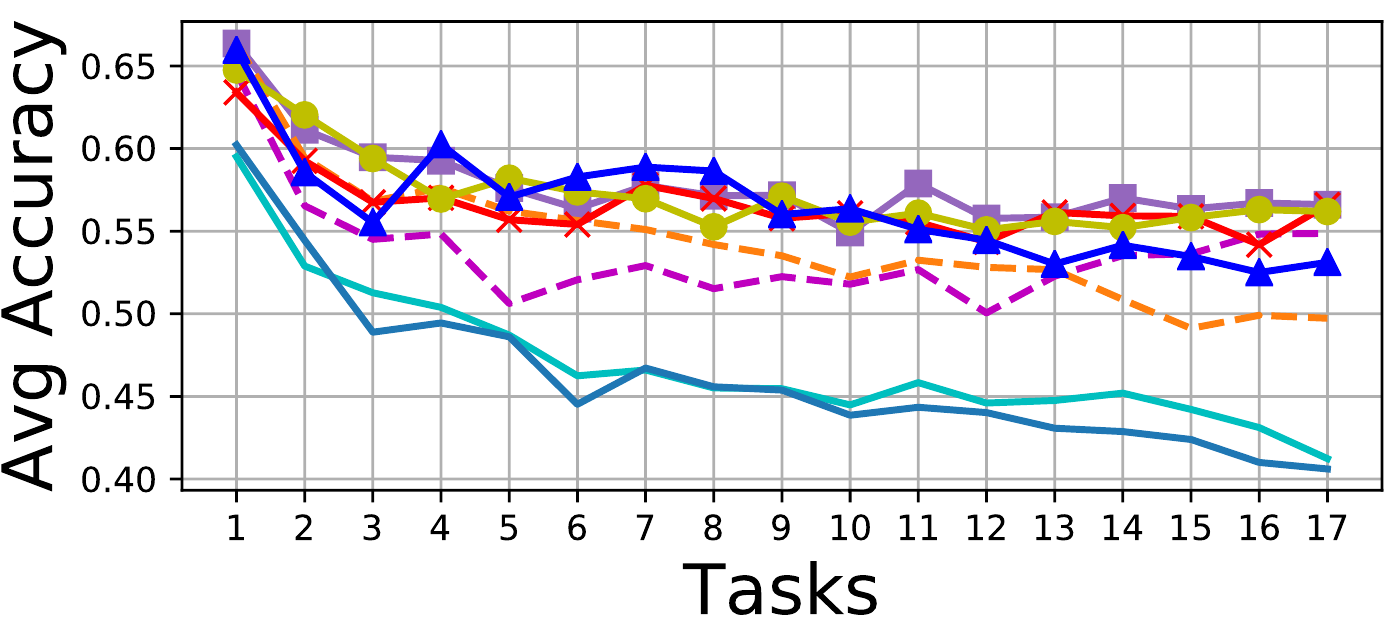}
                \caption{\small 1 Sample}
        \end{center}
        \end{subfigure}
        \begin{subfigure}{0.45\linewidth}
        \begin{center}
          \includegraphics[scale=0.42]{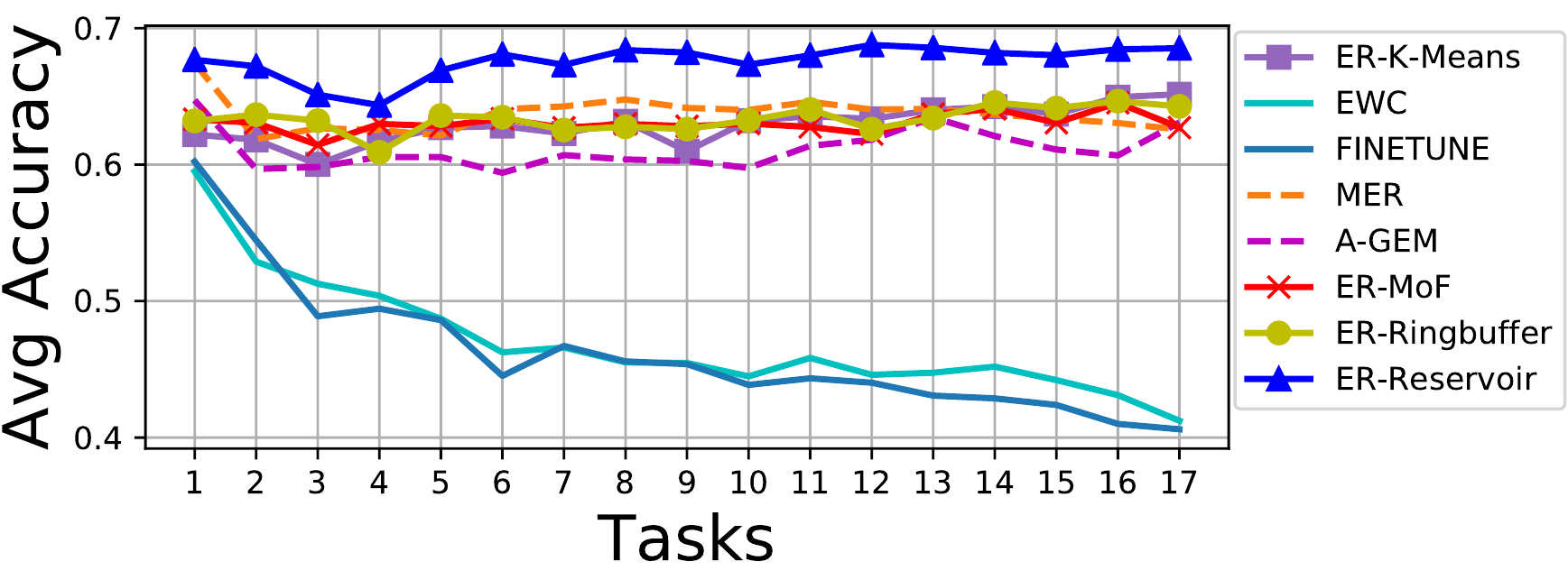}
                \caption{\small 13 Samples}
        \end{center}
        \end{subfigure}
        \end{center}
\caption{\em \small CIFAR: Evolution of average accuracy ($A_k$) as new tasks are learned when using `1' and `13' samples per class. The performance is averaged over $5$ runs. Uncertainty estimates are provided in Tabs~\ref{tab:main_mnist_comp}, \ref{tab:main_cifar_comp}, \ref{tab:main_imagenet_comp}, \ref{tab:main_cub_comp}.}
        \label{fig:cifar_average_accuracy}
\end{figure}

\begin{figure}[!th]
    \begin{center}
        \begin{subfigure}{0.45\linewidth}
        \begin{center}
                \includegraphics[scale=0.42]{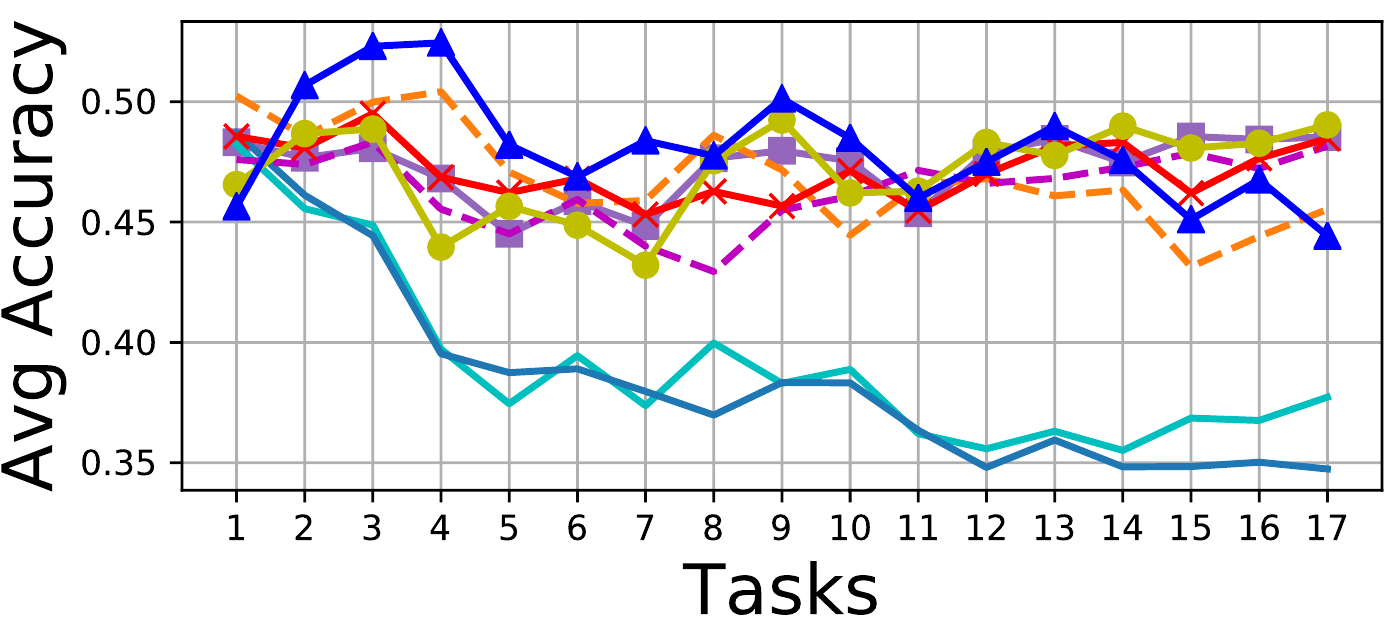}
                \caption{\small 1 Sample}
        \end{center}
        \end{subfigure}%
        \begin{subfigure}{0.45\linewidth}
        \begin{center}
                \includegraphics[scale=0.42]{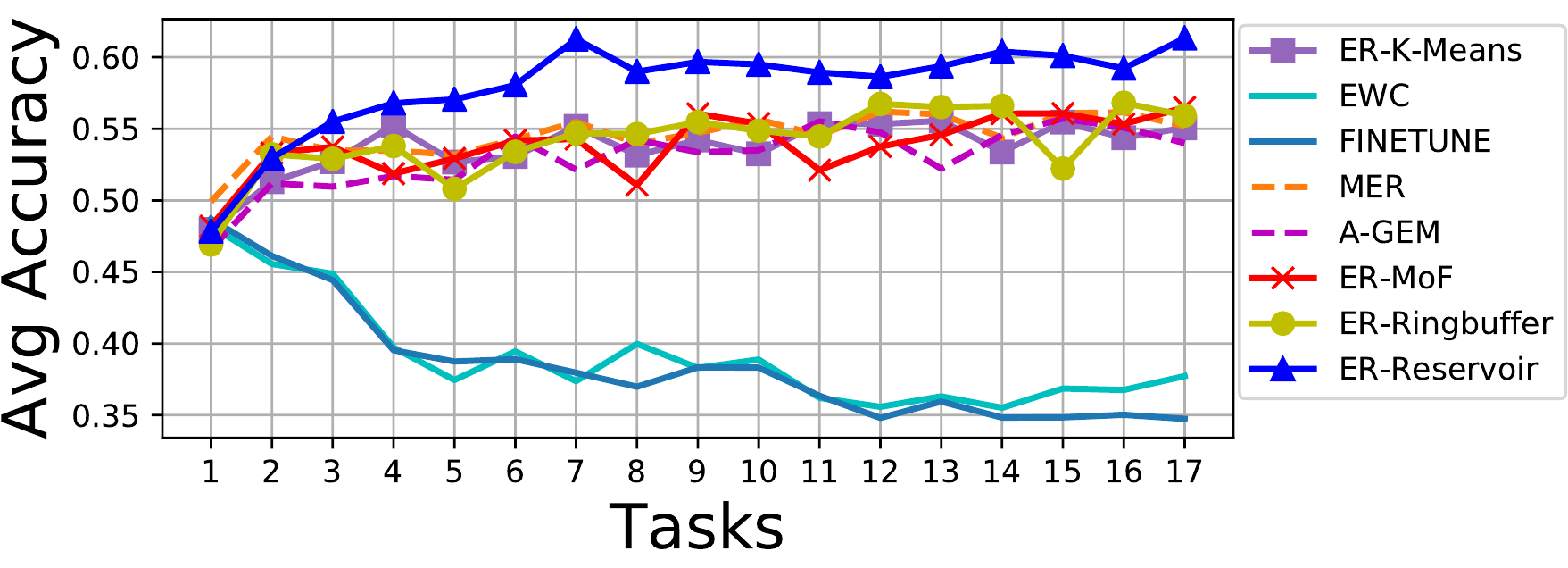}
                \caption{\small 13 Samples}
        \end{center}
        \end{subfigure}
        \end{center}

\caption{\em miniImageNet: Evolution of average accuracy ($A_k$) as new tasks are learned when `1' and `13' samples per class are used.}
        \label{fig:imagenet_average_accuracy}
\end{figure}

\begin{figure}[!th]
	\begin{subfigure}{0.45\linewidth}
	\begin{center}
		\includegraphics[scale=0.42]{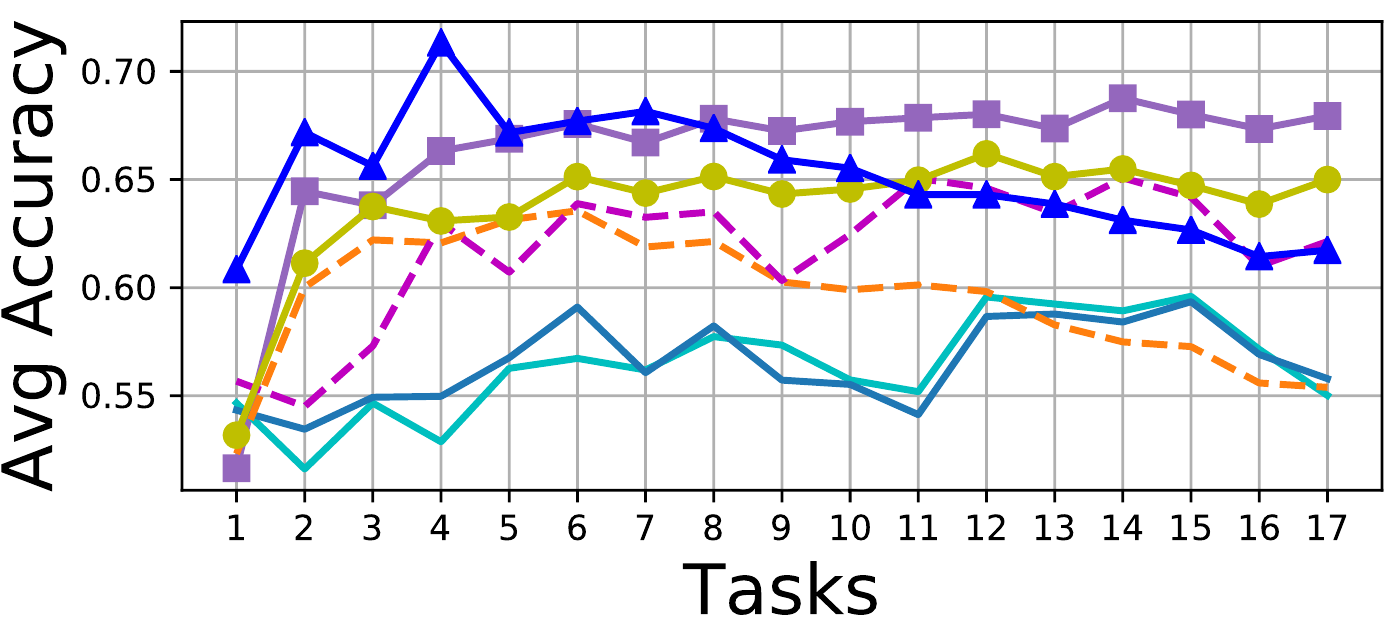}
		\caption{\small 1 Sample}
	\end{center}
	\end{subfigure}%
	\begin{subfigure}{0.45\linewidth}
	\begin{center}
		\includegraphics[scale=0.42]{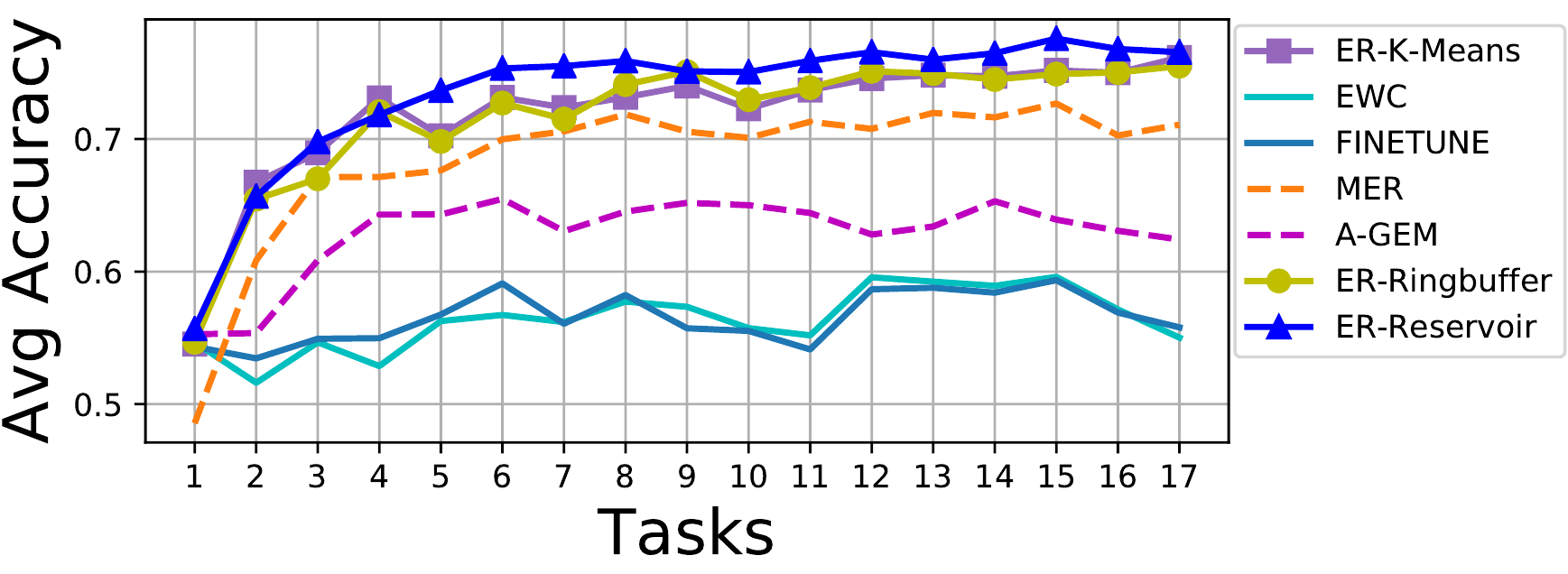}
		\caption{\small 10 Samples}
	\end{center}
	\end{subfigure}%
\caption{\em CUB: Evolution of average accuracy ($A_k$) as new tasks are learned when `1' and `10' samples per class are used.}
	\label{fig:cub_avg_accuracy}
\end{figure}

\section{Further Analysis} \label{sec:further_analysis}

\begin{table}[H]
\centering
\small
\caption{\em \textbf{MNIST Rotation} Performance of task 1 after training on task 2.}
\label{tab:ten_mnist_rotation}
\begin{tabular}{lcccccc|cccccc}
\toprule
\textbf{Task 2 Samples}  & \multicolumn{12}{c}{\textbf{Rotation Angle}} \\
\hline
& \multicolumn{6}{c}{\textbf{10\degree}} & \multicolumn{6}{c}{\textbf{90\degree}} \\
\cmidrule(r){2-13}
& \multicolumn{3}{c}{{\errand}} & \multicolumn{3}{c}{{\agem}} & \multicolumn{3}{c}{{\errand}} & \multicolumn{3}{c}{{\agem}} \\
\cmidrule(r){2-7}  \cmidrule(l){8-13}
& Train & Mem & Test & Train & Mem & Test & Train & Mem & Test & Train & Mem & Test \\
1000   & 85.6 & 1 & 86.2 & 81.5 & 86.6 & 82.5 & 68.7 & 1 & 69.4 & 51.7 & 73.3 & 52.1 \\
20000  & 91.4 & 1 & 91.6 & 91.4 & 1 & 91.5 & 32.7 & 1 & 33.4 & 31.6 & 1 & 33.0 \\
\bottomrule
\end{tabular}           
\end{table}

In Tab.~\ref{tab:ten_mnist_rotation}, we provide train, memory and test set performance on both the {\errand} and {\agem} with two different configurations of tasks; similar tasks ($10$\degree\ rotation), dissimilar tasks ($90$\degree\ rotation). It can be seen from the table, and as argued in the \textsection{\ref{sec:analysis}} of the main paper that {\errand} always achieves the perfect performance on the memory. To achieve the same effect with {\agem} one has to train for more iterations. 

\section{Hyper-parameter Selection} \label{sec:hyper_param}
\begin{table}[!th]
\centering
\small
\caption{\em Hyper-parameters selection on the four benchmark datasets. `lr' is the learning rate, `\textbf{$\lambda$}' is the synaptic strength for {\ewc}, `\textbf{$\gamma$}' is the with in batch meta-learning rate for {\mer}, `s' is current example learning rate multiplier for {\mer}.}
\label{tab:hyper_params}
\begin{tabular}{lcccc}
\toprule
\textbf{Methods}          & \textbf{MNIST}           & \textbf{CIFAR} & \textbf{CUB} & \textbf{miniImageNet} \\
\hline
{\van}                    & lr (0.1)                            & lr (0.03)                        & lr (0.03)                           & lr (0.03) \\
{\ewc}                    & lr (0.1), $\lambda$ (10)            & lr (0.03), $\lambda$ (10)        & lr (0.03), $\lambda$ (10)           & lr (0.03), $\lambda$ (10) \\
{\agem}                   & lr (0.1)                            & lr (0.03)                        & lr (0.03)                           & lr (0.03) \\
{\mer}                    & lr (0.03), $\gamma$ (0.1), s (10)   & lr (0.03), $\gamma$ (0.1), s (5) & lr (0.1), $\gamma$ (0.1), s (5)     & lr (0.03), $\gamma$ (0.1), s (5) \\
{\er}                     & lr (0.1)                            & lr (0.1)                         & lr (0.03)                           & lr (0.1) \\
{\textsc{er-[others]}}    & lr (0.1)                            & lr (0.03)                        & lr (0.03)                           & lr (0.03) \\
\bottomrule
\end{tabular}           
\end{table}

\end{document}